\documentclass[twoside]{article}

\usepackage{PRIMEarxiv}
\usepackage[utf8]{inputenc}
\usepackage[T1]{fontenc}
\usepackage[colorlinks=true, urlcolor=blue, citecolor=black]{hyperref}

\usepackage{todonotes}
\usepackage{graphicx}
\usepackage{caption}
\usepackage{subcaption}
\usepackage{float}
\usepackage{listings}
\usepackage{multirow}
\usepackage{amssymb}
\usepackage{mathtools}
\usepackage{amsmath}
\usepackage{comment}
\usepackage{algorithm}
\usepackage{algpseudocode}
\usepackage{siunitx}
\usepackage{dsfont}
\usepackage{xcolor, colortbl}
\usepackage{tikz}
\usetikzlibrary{positioning}
\usetikzlibrary{shapes.geometric}
\usepackage{rotating}
\usepackage{cite}
\usepackage{enumitem}

\makeatletter
\def\hlineb#1{%
	\noalign{\ifnum0=`}\fi\hrule \@height #1 %
	\futurelet\reserved@a\@xhline}
\makeatother

\title{Global $k$-means\texttt{++}: an effective relaxation of the global $k$-means clustering algorithm}
\author{Georgios Vardakas \\
      Department of Computer Science and Engineering \\
      University of Ioannina \\
      GR 45110, Ioannina, Greece\\
      \texttt{g.vardakas@uoi.gr}
      \And
      Aristidis Likas \\
      Department of Computer Science and Engineering \\
      University of Ioannina, \\
      GR 45110, Ioannina, Greece\\
      \texttt{arly@cs.uoi.gr}
}

\begin{document}
\maketitle              

\begin{abstract}
The $k$-means algorithm is a prevalent clustering method due to its simplicity, effectiveness, and speed. However, its main disadvantage is its high sensitivity to the initial positions of the cluster centers. The global $k$-means is a deterministic algorithm proposed to tackle the random initialization problem of k-means but its well-known that requires high computational cost. It partitions the data to $K$ clusters by solving all $k$-means sub-problems incrementally for all $k=1,\ldots, K$. For each $k$ cluster problem, the method executes the $k$-means algorithm $N$ times, where $N$ is the number of datapoints. In this paper, we propose the \emph{global $k$-means\texttt{++}} clustering algorithm, which is an effective way of acquiring quality clustering solutions akin to those of global $k$-means with a reduced computational load. This is achieved by exploiting the center selection probability that is effectively used in the $k$-means\texttt{++} algorithm. The proposed method has been tested and compared in various benchmark datasets yielding very satisfactory results in terms of clustering quality and execution speed.
\keywords{Clustering \and $k$-means \and clustering error \and global optimization \and global $k$-means \and $k$-means\texttt{++}.}
\end{abstract}
	
\section{Introduction}
Clustering is one of the most important and fundamental tasks in machine learning, data mining, and pattern recognition, with numerous applications in computer science and many other scientific fields~\cite{jain1999data, filippone2008survey, jain2010data}. It is defined as a process of partitioning a set of objects into groups, called clusters so that the data in the same cluster share common characteristics and differ from those in other clusters. While its definition is simple, it is actually a challenging problem to solve. The typical form of clustering is the partitioning of a given dataset $X =\{x_1,\ldots, x_N\}$, $x_i \in R^d$ into $K$ disjoint clusters $\mathcal{C} = \{C_1, \ldots, C_K\}$ so that a specific criterion is optimized. The most widely used optimization criterion is the clustering error. It is defined as the sum of the squared Euclidean distances between each data $x_i \in C_k$ to its cluster center $m_k$ as defined in
\begin{equation}
	  \label{eq:ClusteringError}
	E(C) = \sum_{i=1}^{N} \sum_{k=1}^{K} \textbf{1}_{C_k}(x_i) ||x_i - m_k||^2,
\end{equation}
where $\textbf{1}_{C_k}$ is the indicator function of the set $C_k$, while $M = \{m_1,\ldots, m_K \}$ is the set of $K$ centers computed as the mean of the datapoints of each cluster. The number of ways in which a set of $N$ objects can be partitioned into $K$ non-empty groups is given by Stirling numbers of the second kind:
\begin{equation}
    \label{eq:PossiblePartitions}
	S(N, K) = \frac{1}{K!} \sum_{k=0}^{K} (-1)^{k} (K-k)^N {K \choose k},
\end{equation}
which can be approximated by $K^{N}/K!$ as $N \rightarrow +\infty $~\cite{kaufman2009finding}. It is evident that a complete enumeration of all possible clusterings to determine the global minimum of eq.~\ref{eq:ClusteringError} is computationally prohibitive. It is worth mentioning that this non-convex optimization problem is  $\mathsf{NP}$-hard \cite{cohen2019inapproximability, cohen2021approximability} not only for two clusters ($K=2$)~\cite{aloise2009np}, but also for two-dimensional datasets ($D=2$)~\cite{mahajan2012planar}. The $k$-means algorithm~\cite{macqueen1967some, lloyd1982least} is a widely used method for minimizing clustering error. It is an iterative algorithm that has been extensively utilized in many clustering applications due to its simplicity and speed~\cite{celebi2013comparative}. 

The $k$-means algorithm is computationally fast, but its performance relies heavily on the initialization of the cluster centers, which may lead to a local minimum of the clustering error. Various stochastic global optimization methods have been proposed, such as simulated annealing and genetic algorithms to overcome the above problem, but have not gained wide acceptance. These types of methods involve several hyper-parameters that are difficult to tune, for example, starting temperature, cooling, schedule, population size, and crossover/mutation probability and usually require a large number of iterations, which renders them prohibitive for large datasets. Consequently, stochastic global optimization methods are usually not preferred, but instead, methods with multiple random restarts are commonly used~\cite{jain1999data}. The $k$-means\texttt{++}~\cite{ilprints778} is probably the most widely used center initialization algorithm. It selects the cluster centers by sampling datapoints from a probability distribution that strives to effectively spread them away from each other. Specifically, the algorithm comprises two key steps: i) computing a probability distribution for center selection and ii) sampling a datapoint from this distribution as the initial cluster center. The method iteratively recalculates the distribution in order to select subsequent cluster centers. This iterative process terminates when all $K$ centers have been initialized. 
	
The global $k$-means~\cite{likas2003global} algorithm has also been proposed to effectively solve the $k$-means initialization problem. It constitutes a deterministic global optimization method that does not depend on center initialization. Instead of randomly selecting starting positions for all cluster centers, as is the case with most global clustering algorithms, the global $k$-means proceeds incrementally, attempting to optimally add one new cluster center at each stage $k$ by exploiting the $k-1$ clustering solution. This effective approach is deterministic and does not depend upon any initial conditions or empirically adjustable parameters. Nonetheless, a notable limitation of this method is its high computational cost. It has been empirically proven that both global $k$-means and $k$-means\texttt{++} methods significantly surpass the performance of standard $k$-means~\cite{agrawal2013global, franti2019much}.

It is important to note that the performance of random initialization algorithms, such as $k$-means\texttt{++}, decays as $K$ grows compared to those of the global $k$-means algorithm. 
Large values of $K$ are frequently utilized in clustering scenarios, where the objective is to capture finer substructures within a dataset. In this context, we define a clustering of $X$ as an overclustering when $K>k^{\star}$, with $k^{\star}$ representing the actual number of clusters present in $X$. In various fields such as speech recognition~\cite{1318476} and computational biology~\cite{saeys2016computational}, overclustering is commonly practiced. For example, in computational biology, overclustering ensures that relevant cell types can be discovered even if an expected population is split into sub-populations. Moreover, overclustering is vital in more sophisticated clustering algorithms as an algorithmic step, particularly when dealing with non-convex data structures~\cite{wei2022skeleton, nie2019k}. In overclustering methods, obtaining solutions from the global k-means clustering algorithm is desirable. However, due to the computational burden associated with large values of $N$ and $K$, computationally cheaper alternatives are practically utilized. 
    
In this paper, we propose the \emph{global $k$-means\texttt{++}} clustering algorithm, which is an effective way of acquiring clustering solutions of comparable clustering errors to those that global $k$-means produces without its high computational cost. This is achieved by employing the effective center selection probability distribution of the $k$-means\texttt{++} method. The global $k$-means\texttt{++} algorithm is an attempt to retain the effectiveness of the incremental clustering strategy of the global $k$-means while reducing its computational demand using the efficient stochastic center initialization of the $k$-means\texttt{++} algorithm. The proposed algorithm proceeds incrementally by solving all intermediate sub-problems with $k \in \{1, 2,\ldots, K-1\}$ to provide the clustering solution for $K$ clusters. The underlying idea of the proposed method is that an optimal solution for a clustering problem with $K$ clusters can be obtained using a series of local $k$-means searches that are appropriately initialized. More specifically, at each local search, the $k - 1$ cluster centers are always initially placed at their optimal positions corresponding to the clustering problem with $k-1$ clusters. However, the remaining $k$-th cluster center is placed at several starting positions within the data space that are randomly selected by sampling from the $k$-means\texttt{++} probability distribution. 
    
The rest of the paper is organized as follows. At first, in Section~\ref{sec:RelatedWork}, we give a brief overview of the $k$-means algorithm, the initialization method of the $k$-means\texttt{++}, the global $k$-means clustering algorithm and its variations. Then, in Section~\ref{sec:method}, we present the proposed global $k$-means\texttt{++} clustering algorithm. Finally, in Section~\ref{sec:experiments}, we provide extensive experimental results and comparisons, while in Section~\ref{sec:Conclusion}, we provide our conclusions and future work suggestions.
 
\section{Related work}
\label{sec:RelatedWork}
The \emph{$k$-means} algorithm finds locally optimal solutions with respect to the clustering error (eq.~\ref{eq:ClusteringError}). It is an iterative center-based clustering algorithm that starts with $K$ cluster centers. In its purest form, the $K$ centers $M_k$ are typically selected uniformly at random from the set of datapoints. Then the two-step algorithmic procedure follows iteratively until convergence, which includes the data assignment and the center optimization steps. At the assignment step, every data point $x_i$ is assigned to the cluster $C_j$ with the nearest center $m_j$ (eq.~\ref{eq:assigment}). At the optimization step, each center is updated to the mean of all datapoints assigned to its cluster according to the (eq.~\ref{eq:optimization}) (Lloyd's algorithm~\cite{lloyd1982least}). 
\begin{equation}
    \label{eq:assigment}
     j =\arg\min \limits_k || x_i - m_k ||^2
\end{equation}
\begin{equation}
    \label{eq:optimization}
    m_j = \frac{1}{|C_j|} \sum\limits_{x_i \in C_j} x_i
\end{equation}
This simple optimization methodology is proven to be computationally fast and effective. The main disadvantage of the standard k-means algorithm is that it is a local optimization method with high sensitivity to the original starting positions of the cluster centers. A poor center selection usually leads to local minima of the clustering error. Therefore, to obtain near-optimal solutions using the k-means algorithm, several runs must be scheduled differing in the starting positions of the cluster centers. Among these runs, the solution with the lowest clustering error is naturally chosen and retained.

The \emph{$k$-means\texttt{++}} algorithm~\cite{ilprints778} is considered the most successful method for choosing the initial positions of the cluster centers. 
It randomly selects the center positions from the datapoints using a multinomial probability distribution that strives to effectively distribute the centers away from each other. Specifically, the method begins by choosing the first center position through uniform random selection from the dataset $X$. Then, it computes the distance $d_i$ of each datapoint $x_i$ from its nearest center (eq.~\ref{eq:kmppdistance}) to form the probability vector $P$ (eq.~\ref{eq:kmppdistribution}), and it samples the next center position $m$ from the data using the multinomial distribution with probability vector $P$. This two-step procedure is repeated until the number of centers equals to $K$. After the $k$-means\texttt{++} center initialization procedure, the standard $k$-means algorithm proceeds with the assignment and optimization steps, denoted as (eq.~\ref{eq:assigment}) and (eq.~\ref{eq:optimization}) respectively. These steps are iteratively applied until convergence is achieved.

\begin{equation}
    \label{eq:kmppdistance}
    d_i = \min_{j}||x_i - m_j||^2 \text{, where } i = 1, \ldots, N
\end{equation}
\begin{equation}
    \label{eq:kmppdistribution}
    P = \left(p_1, \ldots, p_N\right) \text{, where } p_i = Pr(m = x_i) = d_i \bigg/ \sum\limits_{j=1}^N d_j
\end{equation}
Choosing centers using the $k$-means\texttt{++} algorithm is computationally fast, while it also guarantees a $\mathcal{O}(\log{k})$-competitive clustering solution~\cite{ilprints778}. 
    
The \emph{global $k$-means} clustering algorithm~\cite{likas2003global} is a deterministic global optimization method that does not depend on any predefined parameter values and employs the $k$-means algorithm as a local search procedure. Instead of randomly selecting initial values for the cluster centers, the global $k$-means algorithm incrementally adds one new cluster center at each stage in an attempt to be optimally placed. To accomplish that, the global $k$-means algorithm proceeds to solve a clustering problem with $K$ clusters by sequentially solving every intermediate
sub-problem with $k$ clusters ($k \in \{1,\ldots, K\}$). In order to solve the problem with $k$ clusters, the global $k$-means utilizes the solution with $k-1$ cluster centers. Specifically, it starts by solving the $1$-means problem at which the optimal position corresponds to the center of the dataset $X$. Then, it solves the $2$-means problem by performing $N$ executions of the k-means algorithm. In each execution $n$, the first cluster center is always initialized at the optimal solution of the $1$-means sub-problem, while the second center is initially set at the data point $x_n$ $(n \in \{1,\ldots, N\})$. The best solution (with the lowest clustering error) obtained after the $N$ executions of the k-means algorithm is considered the solution for the $2$-means clustering sub-problem. Following the same procedure, the solution for $k$ clusters is obtained. Specifically, the procedure begins with the initialization of the $k-1$ centers at the center positions provided by the solution of the $(k-1)$ problem; then, the new $k$-th center is initialized at each $x_n$. The $k$-means algorithm is executed $N$ times while retaining the best clustering solution. The global $k$-means algorithm is very successful in obtaining near-optimal solutions, but computationally expensive for large $N$ as it involves $\mathcal{O}(NK)$ executions of $k$-means on the entire dataset. The complete global $k$-means method is presented in algorithm~\ref{alg:global_kmeans}.
    
\begin{algorithm}[h]
    \caption{The global $k$-means~\cite{likas2003global}}
    \label{alg:global_kmeans}
    \hspace*{\algorithmicindent} \textbf{Input} Dataset $X = \{x_1, \ldots, x_N\}$, Number of clusters $K$. \\
    \hspace*{\algorithmicindent} \textbf{Output} $\mathcal{C}_k$ clustering solutions and $M_k$ cluster centers for every $k \in \{ 1,\ldots,K\}$.
    \begin{algorithmic}[1]
    	\State $m_1 \leftarrow \frac{1}{|X|} \sum\limits_{x_i \in X} x_i$; $M_1 \leftarrow \{m_1\}$.
    	\For {$k = 2, \ldots, K$}
        	\ForAll {$x_i \in X$}
        		\State Run $k$-means with initial centers positions $\{M_{k-1}\}\cup \{x_i\}$ 
                \State Compute the clustering error $E_i(C_k)$ (eq.~\ref{eq:ClusteringError}).
        	\EndFor
    	\State $\{C_k,M_k\} \leftarrow $ Partition and cluster centers of the clustering solution with the minimum clustering error among the $N$ $k$-means runs.
        \EndFor
    \end{algorithmic}
\end{algorithm}

The global $k$-means algorithm is widely acknowledged for its optimization capabilities. However, its prominent drawback lies in its high computational complexity, as each $k$ clustering sub-problem necessitates $\mathcal{O}(N)$ $k$-means executions. Consequently, the practical application of the algorithm is predominantly confined to small datasets. This limitation has prompted researchers to explore heuristic techniques for selecting initial center candidates, with the dual objective of reducing computational complexity while maintaining clustering results of comparable quality to those obtained with global $k$-means. Nonetheless, it is essential to acknowledge that reducing the computational complexity of the global $k$-means algorithm results in the loss of its deterministic nature.

The fast global $k$-means algorithm (FGKM) ~\cite{likas2003global} constitutes an effort to accelerate the global $k$-means. Unlike the original global $k$-means, which necessitates $N$ $k$-means executions for each $k$ sub-problem, the FGKM algorithm performs only a single $k$-means execution. 
In particular, for each value of $k$, the FGKM algorithm computes an upper bound on the clustering error $E_n$ that would arise if the new center were initialized at the specific point $x_n$. After selecting the new center position, the k-means is executed until convergence. The upper bound is defined as $E_n \leq E - b_n$, where $E$ is the error already found for the $k-1$ clustering sub-problem and $b_n$ is given by equation~\ref{eq:bn}. In algorithm~\ref{alg:fast_global_kmeans} we present the FGKM clustering method.
\begin{equation}
	\label{eq:bn}
	b_n = \sum_{j=1}^{N} \max (d^{j}_{k-1} - ||x_n - x_j||^2, 0) 
\end{equation}
\begin{equation}
	\label{eq:i}
	x_i=\arg\max_n b_n 
\end{equation}
The initial position of the $k$ cluster center is set to the data point with minimum $E_n$ or, equivalently, with maximum $b_n$. In equation \ref{eq:bn}, the $d^{j}_{k-1}$ is defined as the squared euclidean distance between $x_j$ and its closest cluster center among the $k-1$ centers obtained so far.
	
\begin{algorithm}[h]
    \caption{The fast global $k$-means~\cite{likas2003global}}
    \label{alg:fast_global_kmeans}
    \hspace*{\algorithmicindent} \textbf{Input} Dataset $X = \{x_1, \ldots, x_N\}$, Number of clusters $K$. \\
    \hspace*{\algorithmicindent} \textbf{Output} $\mathcal{C}_k$ clustering solutions and $M_k$ cluster centers for every $k \in \{ 1,\ldots,K\}$.
    \begin{algorithmic}[1]
    	\State $m_1 \leftarrow \frac{1}{|X|} \sum\limits_{x_i \in X} x_i$; $M_1 \leftarrow \{m_1\}$.
    	\For {$k = 2, \ldots, K$}
    		\State Compute $b_n$ as shown in equation \ref{eq:bn}.
    		\State Select as new initial candidate center position the $i$-th data point as shown in equation \ref{eq:i}.
        	\State Run $k$-means with initial centers positions $\{M_{k-1}\}\cup \{x_i\}$.
        \EndFor
    \end{algorithmic}
\end{algorithm}
	
Several methods have been proposed that modify the global $k$-means or the fast global $k$-means to make it more efficient~\cite{lai2010fast, bagirov2006modified, agrawal2013global}. The efficient global $k$-means clustering algorithm~\cite{xie2011efficient} defines a normalized density function criterion for selecting the top candidate for the new cluster center. However, it demands all pairwise distances $d(x_i, x_j)$ as the FGKM, thus it needs $\mathcal{O}(N^2)$ distance calculations and extra memory space. The fast modified global $k$-means algorithm (FMGKM)~\cite{bagirov2011fast} defines a more sophisticated auxiliary function criterion for selecting the candidates compared to FGKM. Its main drawback is computational complexity because, at each iteration, it requires the whole affinity matrix computation. 
In addition, the FMGKM algorithm exhibits a secondary limitation due to the utilization of an auxiliary function criterion, which introduces a non-convex optimization problem. The fast global $k$-means clustering based on local geometric information~\cite{bai2013fast} is a successful attempt to reduce the computational complexity of FGKM~\cite{likas2003global} and FMGKM~\cite{bagirov2011fast}. The algorithm uses local geometric structure information to decrease the distance computations required in each iteration. It is crucial to emphasize that all related methods exhibit comparable clustering performance to FGKM rather than global $k$-means.

\section{The global $k$-means\texttt{++} algorithm}
\label{sec:method}

\begin{figure}[t]
    \centering 
    \begin{subfigure}[b]{0.49\textwidth}
        \centering
        \includegraphics[width=\textwidth]{"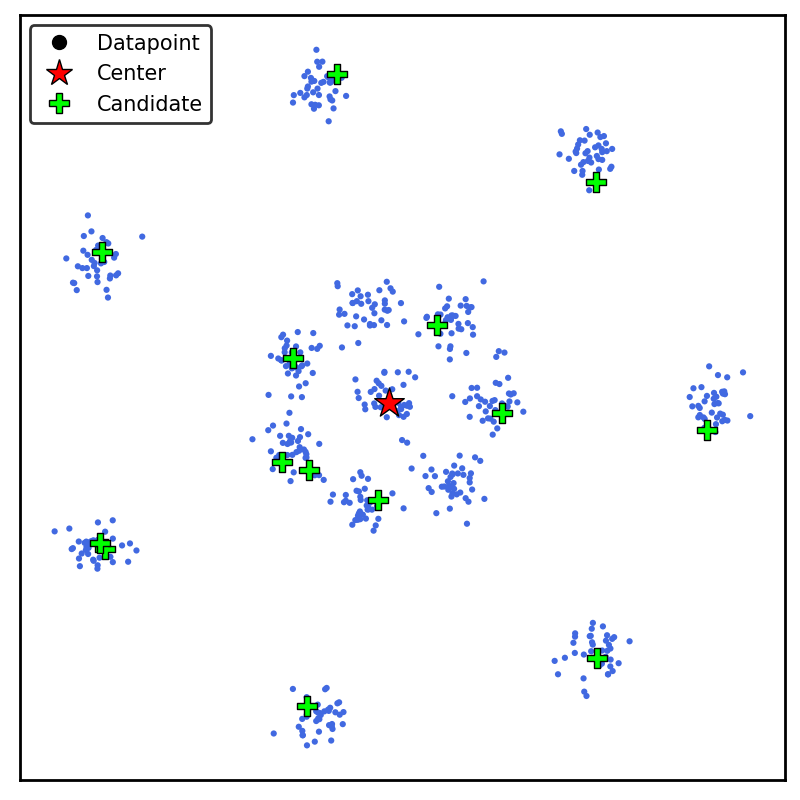"}
        \caption{1-means sub-problem.}
        \label{subfig:1means}
    \end{subfigure}
    \hspace{0em}
    \begin{subfigure}[b]{0.49\textwidth}
        \centering
        \includegraphics[width=\textwidth]{"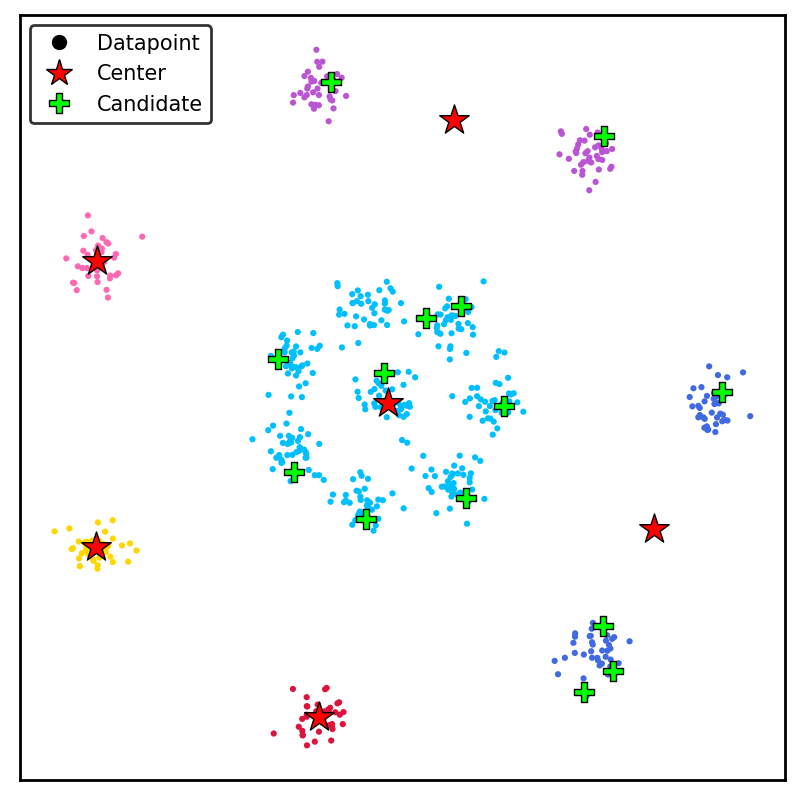"}
        \caption{6-means sub-problem.}
        \label{subfig:7means}
    \end{subfigure}
    \begin{subfigure}[b]{0.49\textwidth}
        \centering
        \includegraphics[width=\textwidth]{"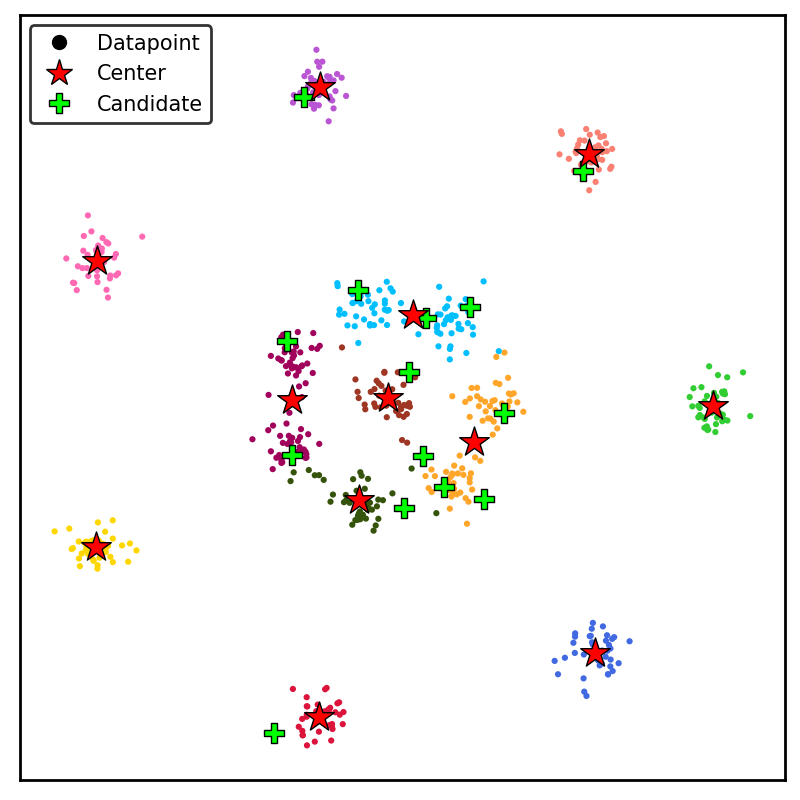"}
        \caption{12-means sub-problem.}
        \label{subfig:12means}
    \end{subfigure}
    \begin{subfigure}[b]{0.49\textwidth}
        \centering
        \includegraphics[width=\textwidth]{"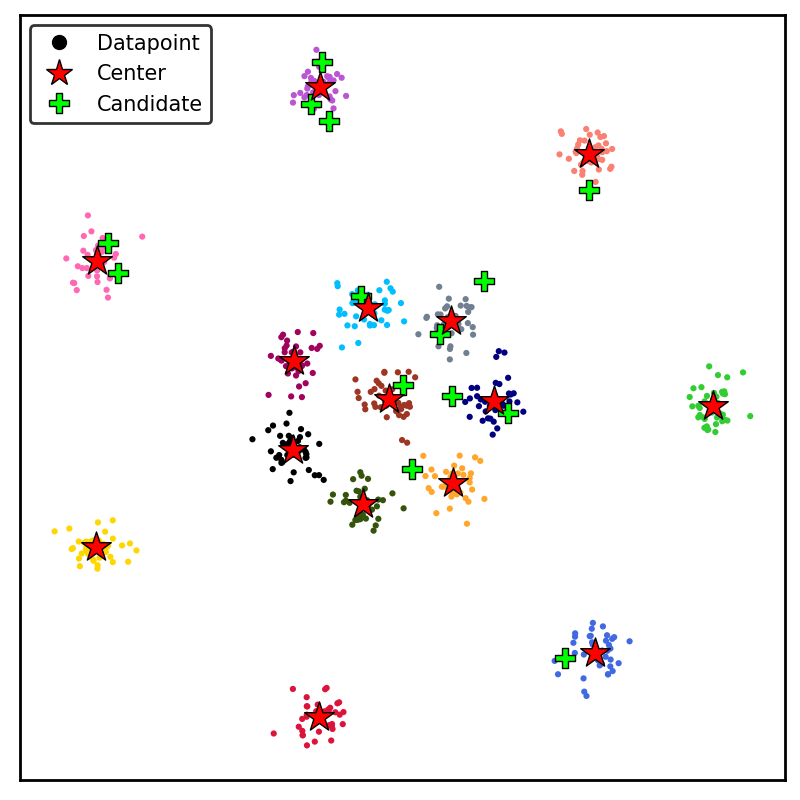"}
        \caption{15-means sub-problem.}
        \label{subfig:15means}
    \end{subfigure}
    \caption{Illustration of a running instance of the algorithm applied to the ``R15'' dataset \cite{veenman2002maximum}. Circles denote the datapoints, the cluster centers are represented by red stars, while the center candidates are marked with green crosses.}    
    \label{fig:Example}
\end{figure}

The global $k$-means is a deterministic algorithm proposed to tackle the random initialization problem but it is computationally expensive. It partitions the data to $K$ clusters by solving all $k$ cluster sub-problems sequentially $k \in \{1,\ldots, K\}$. 
In order to solve the $k$ clustering sub-problem, the method employs a strategy wherein the $k$-means algorithm is executed $N$ times. Each algorithmic cycle considers all $N$ datapoints as potential candidate positions for the new cluster center. This comprehensive exploration of candidate positions aims to identify the optimal center placement for the new cluster center and drastically improves the clustering performance.

An approach to reduce the complexity of the global $k$-means algorithm is to consider a set of $L$ datapoints as candidate positions for the new cluster center, where $L \ll N$. However, an effective candidate center selection strategy is required. Employing a random uniform selection method for the $L$ candidates is not expected to be an viable choice. An ideal candidate selection method should meet the following requirements:
\begin{description}
    \item \emph{Efficiency requirements}
    \begin{enumerate}[label=\roman*.]
        \item Low computational complexity.
        \item Low space complexity.
    \end{enumerate}
    \item \emph{Effectiveness requirement}
     \begin{enumerate}[label=\roman*.]
       \item Selection of a set of high-quality center candidates. 
    \end{enumerate}
\end{description}
We consider as ``high-quality'' a set of centers assigned to seperate data regions. In order to select such center candidates, prioritizing sampling from regions without cluster centers is crucial. Such strategy enhances the likelihood of choosing candidates that belong to different clusters while minimizing the risk of selecting redundant or sub-optimal center positions. For this reason, we have employed the $k$-means\texttt{++} probability distribution as the candidate selection method. The effectiveness of this distribution in the initial placement of cluster centers makes it an excellent choice for addressing the aforementioned requirements.

\begin{algorithm}[h]
    \caption{\texttt{Global $k$-means++}}
    \label{alg:global_kmeans_pp}
    \hspace*{\algorithmicindent} \textbf{Input} Dataset $X = \{x_1, \ldots, x_N\}$, Number of clusters $K$, Number of candidates $L$, Sampling method $S$. \\
    \hspace*{\algorithmicindent} \textbf{Output} $\mathcal{C}_k$ clustering solutions and $M_k$ cluster centers for every $k \in \{ 1,\ldots,K\}$. 
    \begin{algorithmic}[1]
    	\State $m_1 \leftarrow \frac{1}{|X|} \sum\limits_{x_i \in X} x_i$; $M_1 \leftarrow \{m_1\}$; $D \leftarrow \left(||x_1 - m_1||^{2}, \ldots, ||x_N - m_1||^{2}\right)$. \Comment{Optimal initialization for $k=1$.}
    	\For {$k = 2, \ldots, K$}
            \ForAll {$x_i \in X$}  \Comment{Precomputed by the $(k-1)$-means execution.}
        		\State $d_i \leftarrow \min\limits_{m_j \in M_{k-1}} ||x_i - m_j||^{2}$.
                \State $D[i] \leftarrow d_i$
        	\EndFor
            
            \If{$S=\texttt{Batch}$} \Comment{Batch sampling strategy.}
                \State $\{c_1, \ldots, c_L\} \leftarrow$ \texttt{Batch Sampling}$\left(X, M_{k-1}, L, D\right)$.
            \Else   \Comment{Sequential sampling strategy.}
                \State $\{c_1, \ldots, c_L\} \leftarrow$ \texttt{Sequential Sampling}$\left(X, M_{k-1}, L, D\right)$.
            \EndIf
    	\ForAll {$c_\ell \in \{c_1, \ldots, c_L\}$} \Comment{$L$ $k$-means executions.}
    		\State Run $k$-means with initial centers positions $M_{k-1}\cup\{c_\ell\}$.
            \State Compute the clustering error $E_i\left(C_k\right)$ (eq.~\ref{eq:ClusteringError}).
    	\EndFor
    	\State $\{C_k,M_k\} \leftarrow $ Partition and cluster centers of the clustering solution with the minimum clustering error among the $L$ $k$-means runs.
        \EndFor
    \end{algorithmic}
\end{algorithm}

\begin{algorithm}[h]
    \caption{\texttt{Batch Sampling}}
    \label{alg:BatchSampling}
    \hspace*{\algorithmicindent} \textbf{Input} Dataset $X = \{x_1, \ldots, x_N\}$, Set of cluster centers $M$, Number of candidates $L$, Distance vector $D$.\\
    \hspace*{\algorithmicindent} \textbf{Output} $\{c_1, \ldots, c_L\}$ set of initial center candidates. 
    \begin{algorithmic}[1]
        \State Compute the probability vector $P = \left(p_1, \ldots, p_N\right)$, where $p_i = Pr(m_k = x_i) = {\displaystyle d_i \bigg/\sum\limits_{j=1}^N d_j}$.
        \State $\{c_1, \ldots, c_L\} \leftarrow$ Sample without replacement $L$ center candidates from dataset $X$ by $k$-means\texttt{++} probability vector $P$.
    \end{algorithmic}
\end{algorithm}

\begin{algorithm}[h]
    \caption{\texttt{Sequential Sampling}}
    \label{alg:SequentialSampling}
    \hspace*{\algorithmicindent} \textbf{Input} Dataset $X = \{x_1, \ldots, x_N\}$, Set of cluster centers $M$, Number of candidates $L$, Distance vector $D$.\\
    \hspace*{\algorithmicindent} \textbf{Output} $\{c_1, \ldots, c_L\}$ set of initial center candidates. 
    \begin{algorithmic}[1]
    \For {$\ell = 1, \ldots, L$}
        \State Compute the probability vector $P = (p_1, \ldots, p_N)$, where $p_i = Pr\left(r_k = x_i\right) = {\displaystyle d_i \bigg/\sum\limits_{j=1}^N d_j}$.
        \State $\{c_1, \ldots, c_{\ell-1}\} \cup \{c_\ell\} \leftarrow$ Sample without replacement one $c_{\ell}$ center candidate from dataset $X$ by $k$-means\texttt{++} probability vector $P$.
        \ForAll {$x_i \in X$}
            \State $d_i \leftarrow \min\limits_{r_j \in M_{k-1} \cup \{c_1, \ldots, c_\ell\}} ||x_i - r_j||^{2}$. 
        \EndFor
    \EndFor
    \end{algorithmic}
\end{algorithm}

In this work, we propose the global $k$-means\texttt{++} algorithm, an effective relaxation of the global $k$-means method. Its main difference with the global $k$-means is that the proposed method requires only $L$ executions of $k$-means for each $k$ cluster sub-problem, where $L \ll N$. The primary goal of our proposed clustering method is to provide high-quality results that are comparable to those of the global $k$-means algorithm, while retaining low computational and space complexity. The complete global $k$-means\texttt{++} method is presented in algorithm \ref{alg:global_kmeans_pp}. As with any other global $k$-means relaxation method, in order to reduce the computational requirements, we sacrifice determinism by using an effective stochastic initial center selection procedure. 

Specifically, to solve a clustering problem with $K$ clusters, the method proceeds as follows. Initially, it addresses the $1$-means sub-problem by setting the center $m_1$ as the mean of the entire dataset (step 1 in algorithm \ref{alg:global_kmeans_pp}). In the $k$\textsubscript{th} step, the set of centers is denoted as $M_k$, and the distance vector is represented by $D = [d_1, \ldots, d_N]$, where each $d_i$ indicates the distance of $x_i$ to each center $m_j$. Subsequently, for each $k \in {2, \ldots, K}$, it recursively tackles the $k$ cluster sub-problem by leveraging the solution of the previous $(k-1)$ cluster sub-problem. To solve for $k$ clusters, it utilizes the center positions obtained from the $(k-1)$ sub-problem as initialization for the $k-1$ centers. In order to initialize the new $k$-th center, it utilizes the probability distribution $P$ (eq.~\ref{eq:kmppdistribution}) of the $k$-means\texttt{++} center selection strategy. 

We can sample candidates from probability distribution $P$ through either batch~\ref{alg:BatchSampling} or sequential~\ref{alg:SequentialSampling} sampling. On the one hand, batch sampling does not require any additional computations as it utilizes the precomputed center-to-datapoint distances (step 8 in algorithm~\ref{alg:global_kmeans_pp}). With this method, we can sample a set of $L$ candidate datapoints, denoted as $\{c_1, \ldots, c_L\}$, without replacement, using the probability distribution $P$ (algorithm~\ref{alg:BatchSampling}). On the other hand, sequential sampling (algorithm~\ref{alg:SequentialSampling}) samples one candidate at a time (step 3 in algorithm~\ref{alg:SequentialSampling}) using the distribution $P$. It considers both the converged solutions of the $k-1$ centers and the candidates already sampled (steps 4 to 6 in algorithm~\ref{alg:SequentialSampling}). Although the sequential sampling method incurs the cost of recalculating the minimum distance values, it provides a better spread of the samples.

After obtaining the set of candidates, for each candidate $c_\ell \in \{c_1, \ldots, c_L\}$, the method performs one execution of the $k$-means algorithm (steps 12 to 15 in Algorithm~\ref{alg:global_kmeans_pp}). Subsequently, the best among the $L$ clustering solution is selected based on the minimum clustering error (step 16 in Algorithm~\ref{alg:global_kmeans_pp}). Importantly, it should be noted that steps 12 to 15 of Algorithm~\ref{alg:global_kmeans_pp} can be executed concurrently or in parallel. This parallel execution allows for efficient computation and optimization. Furthermore, it is crucial to highlight that the algorithm provides a clustering solution for every $k\in \{1,\ldots, K\}$.

In Figure~\ref{fig:Example}, we present an illustration of a running example of the algorithm using the 2-dimensional ``R15'' dataset~\cite{veenman2002maximum}. The circles denote the datapoints, while their color indicates the cluster category that they belong to, asserted by the global $k$-means\texttt{++} algorithm. With the red star, we represent the converged cluster centers for each clustering sub-problem $k$, while with the green cross-symbols, we show the initial candidate position of the next cluster center. In each sub-figure~\ref{subfig:1means}-\ref{subfig:15means}, the algorithm solves the current $k$-means sub-problem. Based on the current solution, the algorithm samples the next center candidates from the $k$-means\texttt{++} distribution. The figure demonstrates that the method samples high-quality center candidates.

While the global $k$-means requires $\mathcal{O}(NK)$ $k$-means executions, the global $k$-means\texttt{++} only requires $\mathcal{O}(LK)$, where generally $L \ll N$. Additionally, we have empirically observed that $k$-means converges very fast as $k$ grows. The speed up in convergence is reasonable, since in order to solve each $k$ cluster sub-problem, the method exploits the centers of $k-1$ cluster sub-problem that are already positioned sufficiently well. Therefore the $k$-means does not require many iterations to converge.
	
It should be noted that the proposed method is a relaxation of the global $k$-means algorithm rather than an optimization of the FGKM similar to the methods discussed in the related work. Its simplicity and speed are due to the fact that it does not require additional computation to select the $L$ candidates. For example, the affinity matrix, pairwise distance computation between datapoints, eigenvalues, or sorting algorithms like other methods mentioned in the related work section. The center-to-data point distances required to define the probability vector $P$ have already been computed by the $k$-means procedure, thus the method does not involve any additional distance computations. Note also that solving the $k$-means problem with an incremental procedure has many advantages because, due to the unsupervised nature of the clustering problem, it is usually desirable to obtain solutions for different $k$ that are evaluated using appropriate quality criteria (e.g. silhouette~\cite{rousseeuw1987silhouettes}) for selecting the appropriate number of clusters.
	
\section{Experiments}
\label{sec:experiments}
In this section, we present our experimental study to evaluate the effectiveness of the proposed global $k$-means\texttt{++} clustering method, both in batch (gl\texttt{++} (b)) and sequential sampling (gl\texttt{++} (s)) settings. Our study involved comparisons against other clustering methods, including global $k$-means (gl)~\cite{likas2003global}, FGKM (fgl)~\cite{likas2003global}, $k$-means\texttt{++} ($k$-ms\texttt{++})~\cite{ilprints778}, and standard $k$-means with random uniform initialization (rnd)~\cite{lloyd1982least}.\footnote{Experiments were carried on a machine with an Intel\textsuperscript{\textregistered} Core\textsuperscript{TM} i7-8700 CPU at $\SI{3.20}{\giga\hertz}$ and $\SI{16}{\giga\byte}$ of RAM.} Our implementation of the global $k$-means\texttt{++} clustering algorithm is available in the following GitHub repository: \href{https://github.com/gvardakas/global-kmeans-pp.git}{https://github.com/gvardakas/global-kmeans-pp.git}.

\begin{table}[h]
	\centering
	\caption{Descriptions of datasets.}
	\label{tab:Datasets}
	\begin{tabular}{l@{\quad}c@{\quad}c@{\quad}c}
		\hlineb{1.5pt}
		Dataset & $N$ & $D$ & Source\\
    		\hlineb{1.5pt}
            Breast Cancer & 569 & 30 & \cite{Dua:2019} \\
            MNIST & 60000 & $28 * 28$ & \cite{lecun-mnisthandwrittendigit-2010}\\
            Pendigits & 7494 & 16 & \cite{Dua:2019} \\
            Wine & 178 & 13 & \cite{Dua:2019} \\
    		\hlineb{1.5pt}
	\end{tabular}
\end{table}

\subsection{Datasets}
In order to evaluate the proposed algorithm, a series of experiments have been conducted on various benchmark and publicly available datasets. Moreover, we deliberately selected datasets that exhibited a broad spectrum of data characteristics, encompassing factors such as the number of samples $N$ and the data dimensionality $D$, the complexity, and the domain of origin. The description of the real and synthetic datasets is presented in table \ref{tab:Datasets}.\footnote{The synthetic datasets are available in the following GitHub repository: \href{https://github.com/deric/clustering-benchmark.git}{https://github.com/deric/clustering-benchmark.git}.} As a pre-processing step, we used min-max normalization to map the attributes of each real dataset to the $[0, 1]$ interval to prevent attributes with large ranges from dominating the distance calculations and avoid numerical instabilities in the computations~\cite{milligan1988study}. 
	
\subsection{Evaluation}
The evaluation of clustering methods encompasses various approaches, including internal and external metrics when the ground truth cluster labels are available. However, our investigation focuses solely on treating it as an optimization problem, with the primary objective being the minimization of clustering error. Consequently, we intentionally disregard any class labels in the data, as our emphasis lies not on assessing external clustering metrics or determining the number of clusters. Instead, we focus on clustering error $E(C_k)$ (eq.~\ref{eq:ClusteringError}) as the performance measure for the different methods, enabling direct assessment of their error minimization capabilities. 

Specifically, in order to evaluate the performance of the different methods, we have computed the relative Percentage Error, defined as 
\begin{equation}
    \label{eq:PercentageError}
    PE = \frac{E(C_k) - E(C_{k}^{\star})}{E(C_{k}^{\star})} \times 100\%,
\end{equation}
where $E(C_{k}^{\star})$ is the clustering error of the baseline method (global $k$-means), while $E(C_k)$ is the error provided by each of the compared methods (refer to Fig.~\ref{fig:Breast-Cancer}, \ref{fig:Pendigits} and \ref{fig:Wine}). Considering the Mnist dataset, the global $k$-means did not terminate in reasonable time due to its high computational complexity. In that case, we have used the difference in clustering errors, relative to the best-performing algorithm to facilitate a transparent comparison (see Fig.~\ref{fig:Mnist}). Additionally, we studied the convergence speed of each method for each $k$ by presenting the CPU time required by each algorithm (Table~\ref{tab:CPU_Time}) and the average number of iterations needed for $k$-means to converge (Fig.~\ref{fig:avg_n_iter}). In this way, we provide comprehensive information regarding each method's ability to minimize clustering error as well as its computational speed and efficiency.

\subsection{Experimental Setup}
In our experimental study, we executed the compared methods for a maximum number of clusters, denoted as $K$ and evaluated all clustering solutions for $k\in\{1,\ldots,K\}$ in terms of clustering error. We chose the maximum number of clusters $K$ based on the dataset size, with $K=30$ for smaller datasets (Breast Cancer and Wine) and $K=50$ for medium (Pendigits) and larger datasets (Mnist). For each dataset, we run the compared algorithms with $L$ values of 10, 25, 50, and 100. This range of values allowed us to evaluate the impact of $L$ on the performance of the clustering methods.

In the case of the randomly initialized methods, namely $k$-means\texttt{++} and standard $k$-means, we have defined the parameter $L$ to represent the number of restarts within each sub-problem $k$. In contrast, for the global optimization methods of global $k$-means\texttt{++} and FGKM, only a single execution was performed, considering $L$ candidates. We implemented this approach to ensure a fair comparison with the randomly initialized methods, which do not utilize prior sub-problem solutions in their optimization procedure. It is worth noting that the FGKM, as presented in related work~\cite{likas2003global}, originally considers only one candidate (eq.~\ref{eq:i}). However, in our study, we relaxed this limitation by allowing the method to select the top $L$ candidates in each iteration that maximize equation~\ref{eq:bn}. In the following experiments, we evaluate the optimization capabilities of each method, the average number of iterations required, as well as the time needed for convergence to the clustering solution. In summary for each dataset, we conducted the following experiments:

    
\begin{itemize}
    \item We selected $K=30$ or $K=50$ as the maximum number of clusters depending on the size of the dataset. 
    \item We executed one run of global $k$-means, global $k$-means\texttt{++}, and FGKM. For the two latter methods, we considered $L$ equal to 10, 25, 50 and 100 candidates in each $k$ cluster sub-problem.
    \item The $k$-means\texttt{++} and standard $k$-means methods were initialized $L$ times for each $k=1,\ldots, K$, and we present the solution with lowest clustering error for each $k$ cluster sub-problem.
\end{itemize}
	
\subsection{Results}

    \begin{figure}[t]
        \centering
        \begin{subfigure}[b]{0.49\textwidth}
            \centering
            \includegraphics[width=\textwidth]{"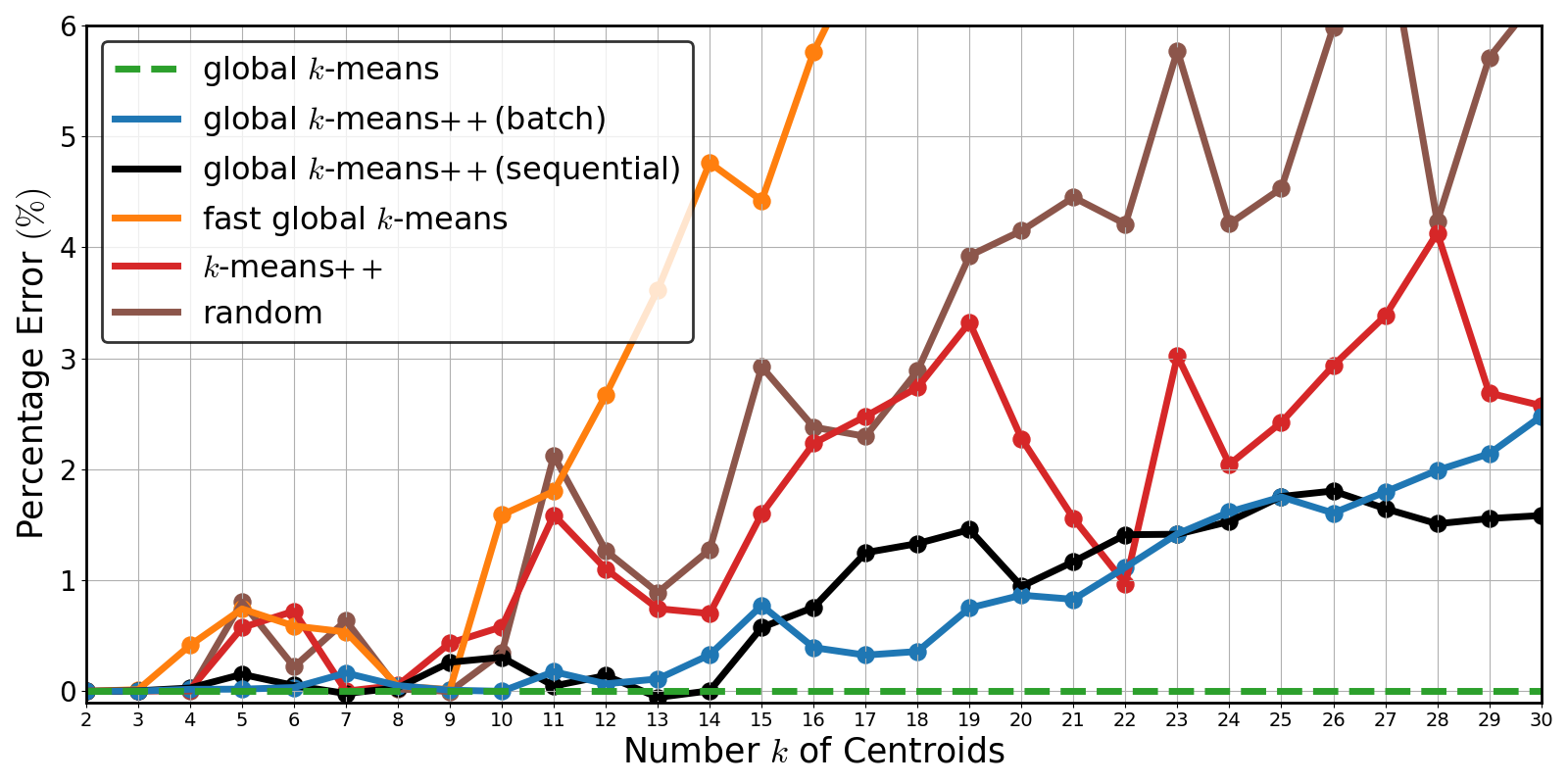"}
            \caption{}
            \label{}
        \end{subfigure}
        \hspace{0em}
        \begin{subfigure}[b]{0.49\textwidth}
            \centering
            \includegraphics[width=\textwidth]{"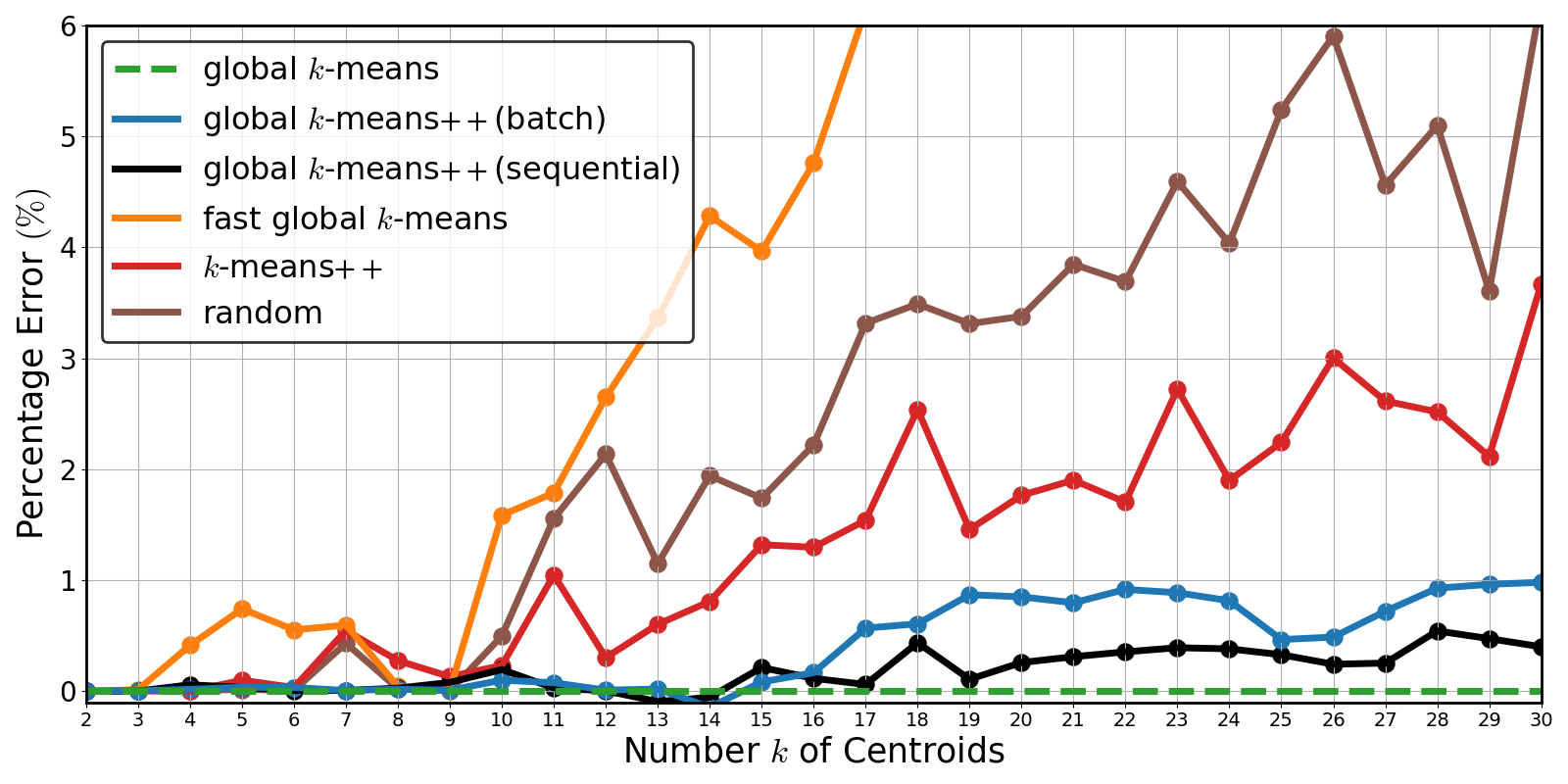"}
            \caption{}
            \label{}
        \end{subfigure}
        \begin{subfigure}[b]{0.49\textwidth}
            \centering
            \includegraphics[width=\textwidth]{"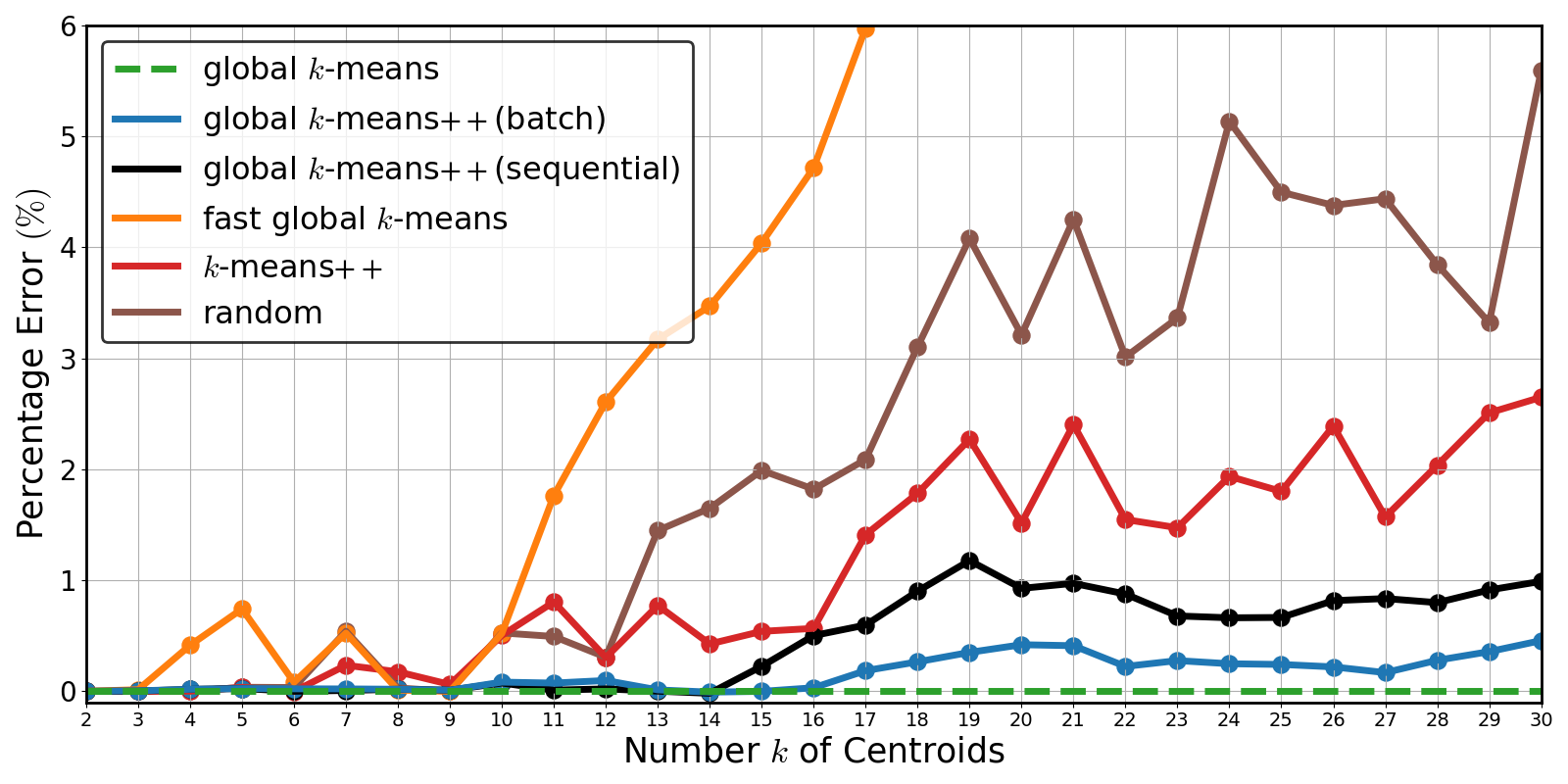"}
            \caption{}
            \label{}
        \end{subfigure}
        \hspace{0em}
         \begin{subfigure}[b]{0.49\textwidth}
            \centering
            \includegraphics[width=\textwidth]{"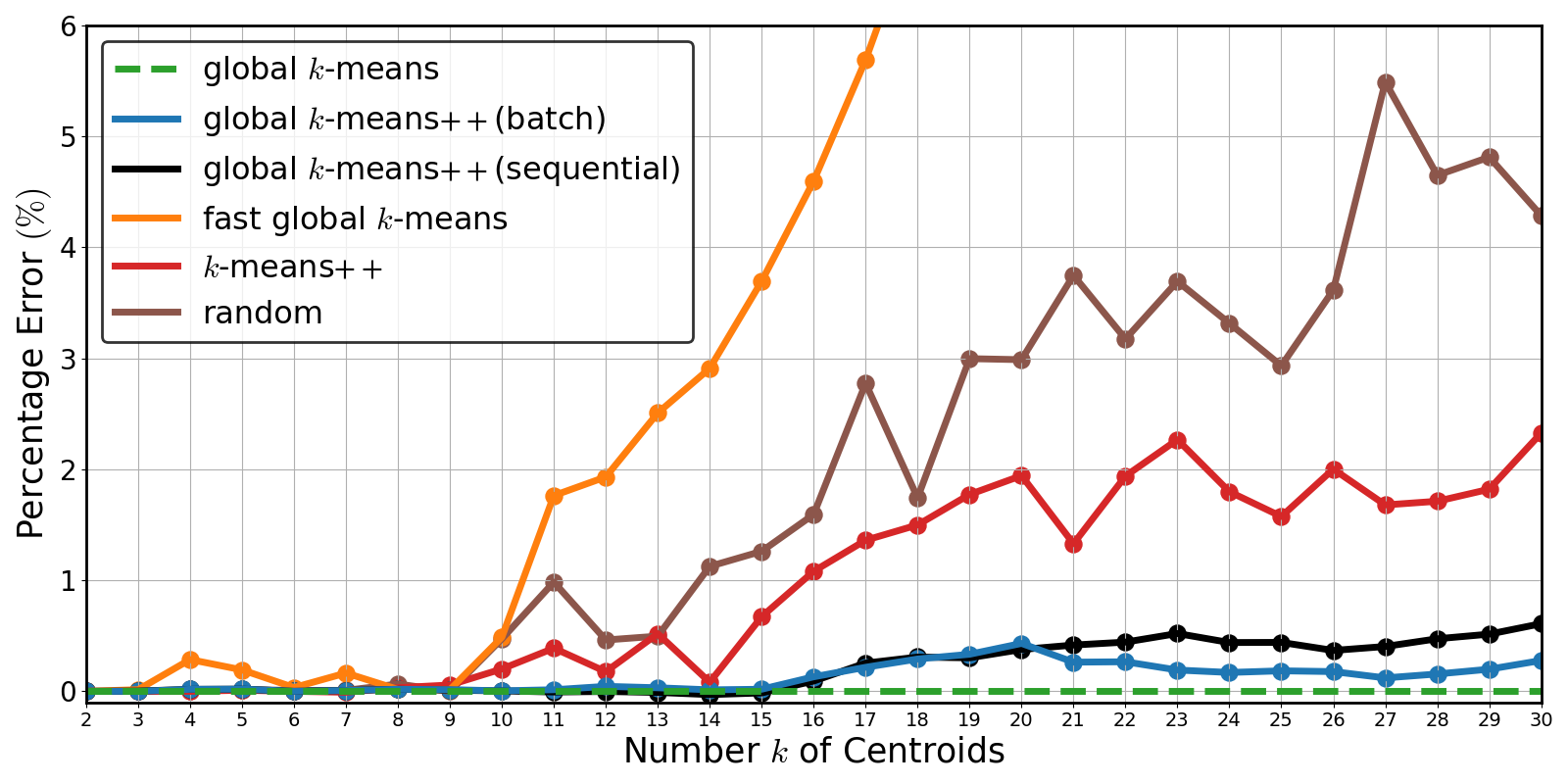"}
            \caption{}
            \label{}
        \end{subfigure}
        \caption{Relative Percentage Error for the Breast Cancer, for different $L$ values. (a) $L=10$ (b) $L=25$ (c) $L=50$ (d) $L=100$.}    
        \label{fig:Breast-Cancer}
    \end{figure}
	
\begin{figure}[t]
    \centering 
    \begin{subfigure}[b]{0.49\textwidth}
        \centering
        \includegraphics[width=\textwidth]{"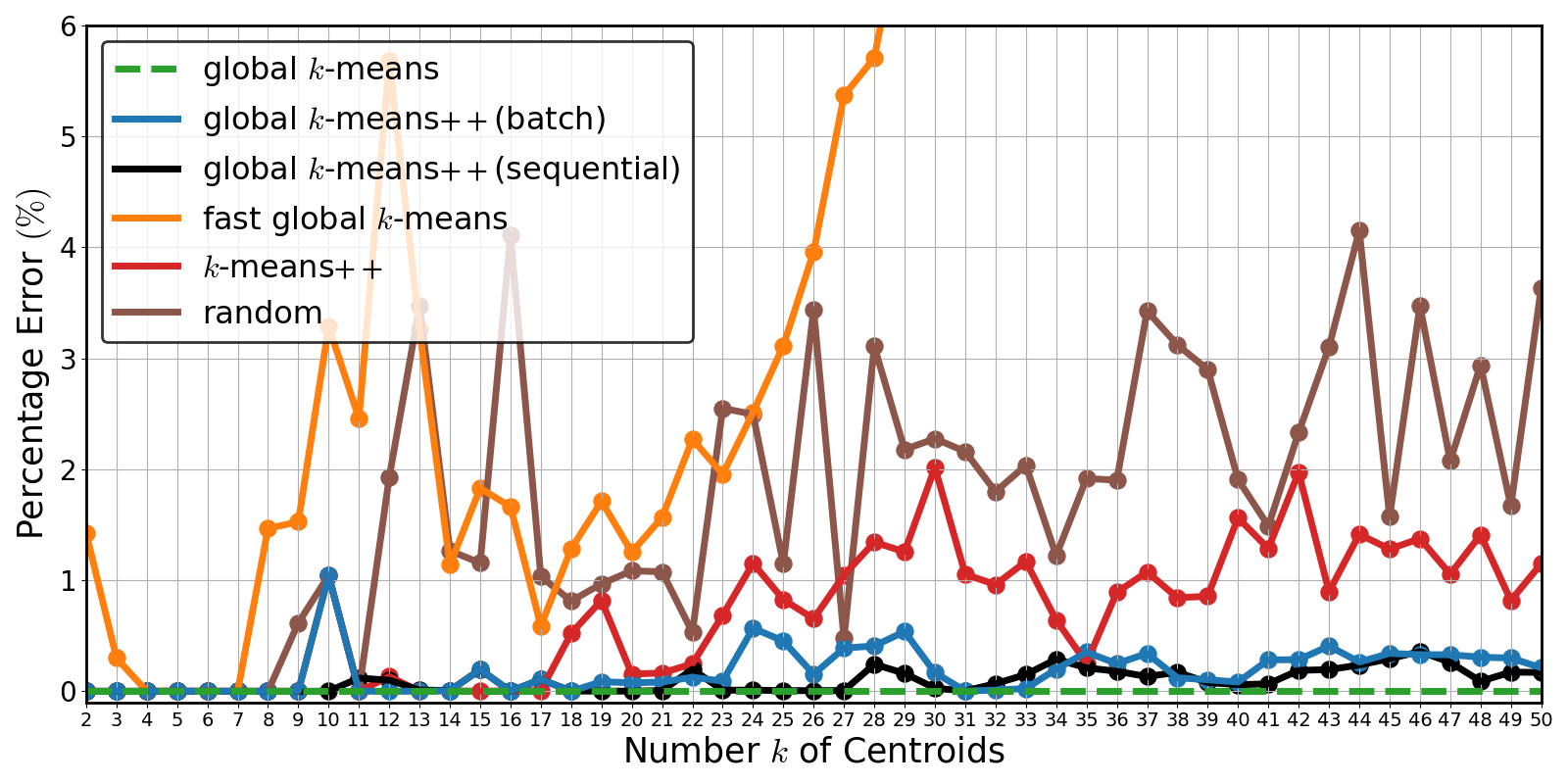"}
        \caption{}
        \label{}
    \end{subfigure}
    \hspace{0em}
    \begin{subfigure}[b]{0.49\textwidth}
        \centering
        \includegraphics[width=\textwidth]{"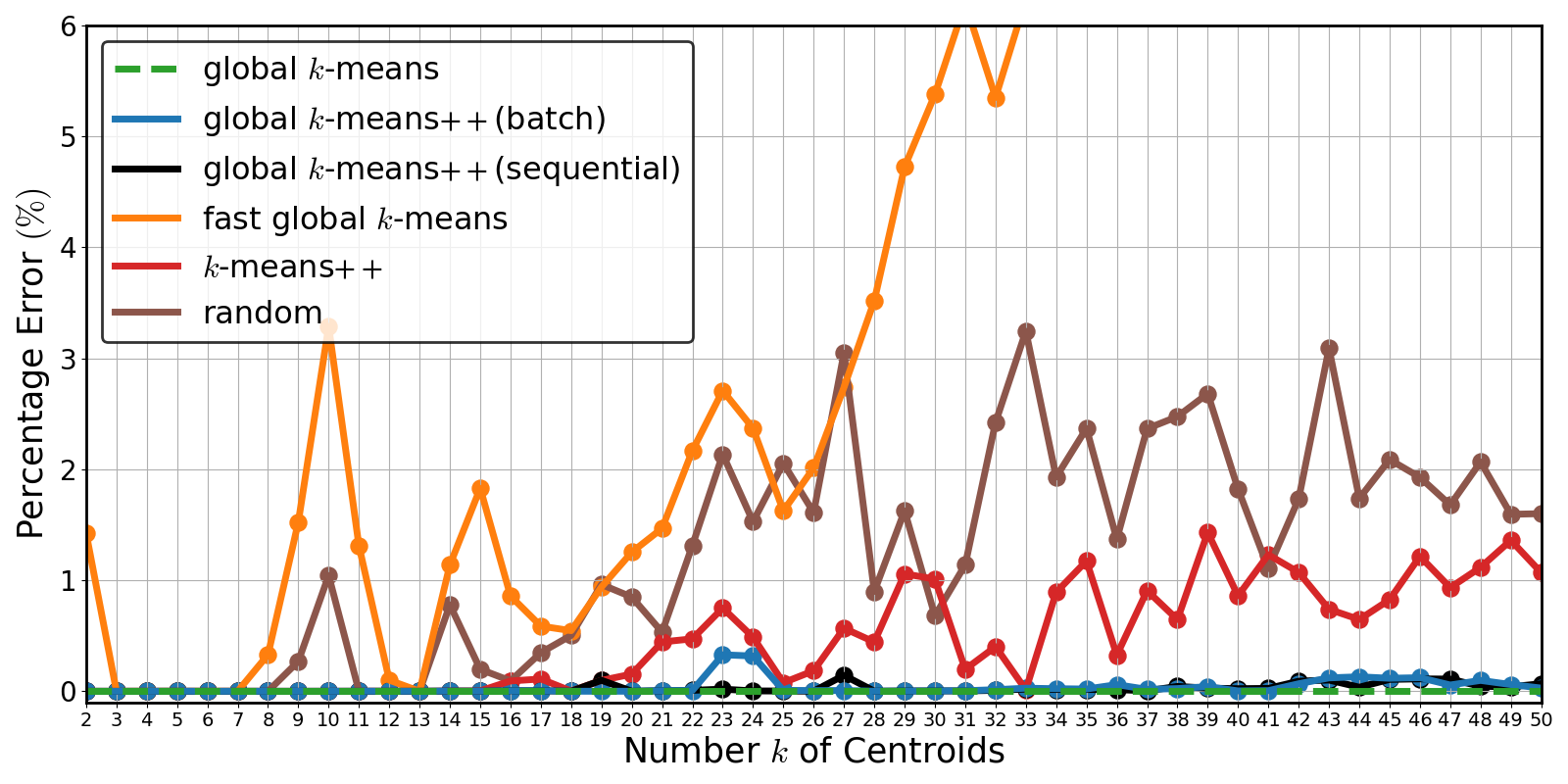"}
        \caption{}
        \label{}
    \end{subfigure}
    \begin{subfigure}[b]{0.49\textwidth}
        \centering
        \includegraphics[width=\textwidth]{"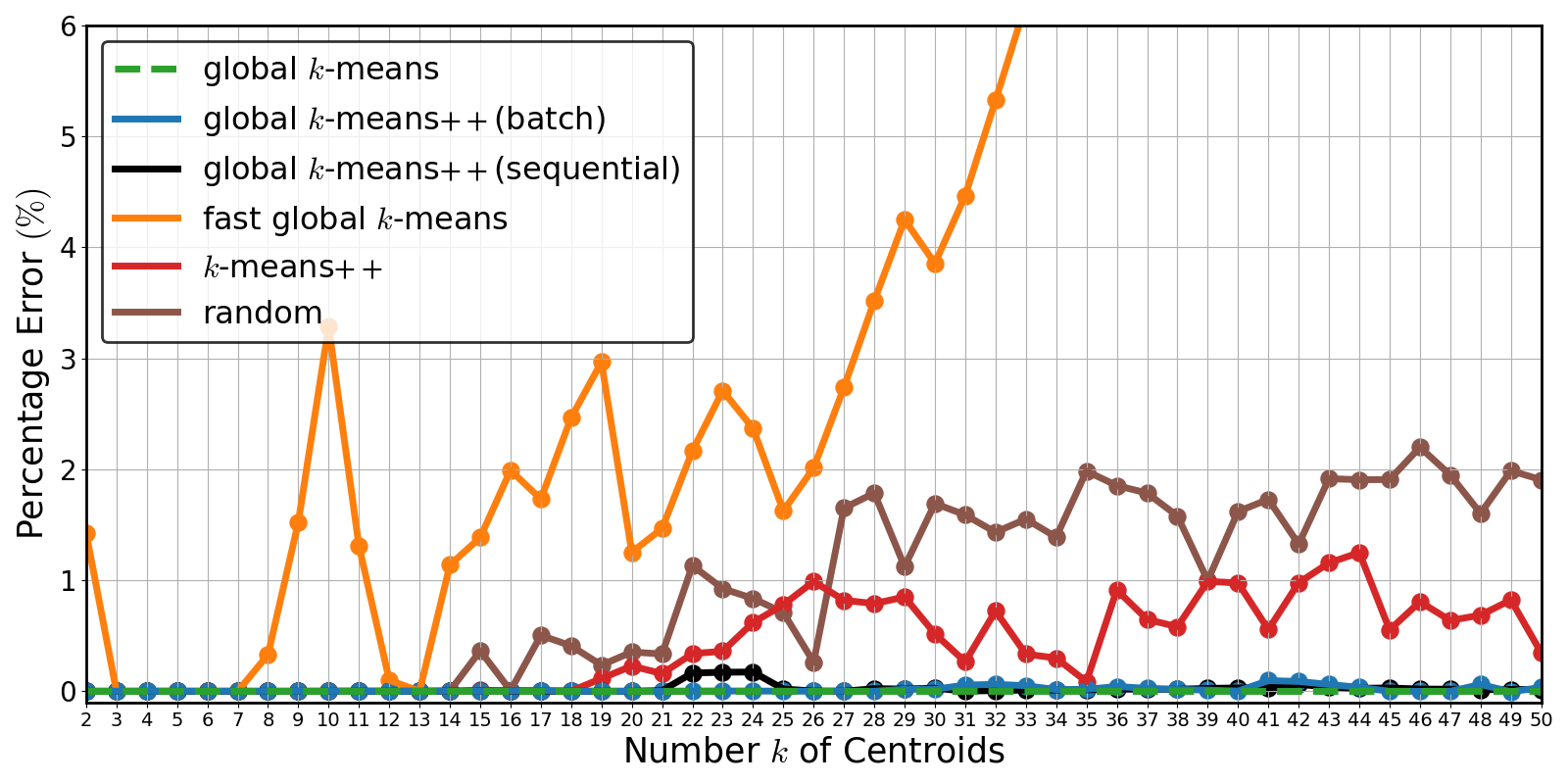"}
        \caption{}
        \label{}
    \end{subfigure}
    \hspace{0em}
    \begin{subfigure}[b]{0.49\textwidth}
        \centering
        \includegraphics[width=\textwidth]{"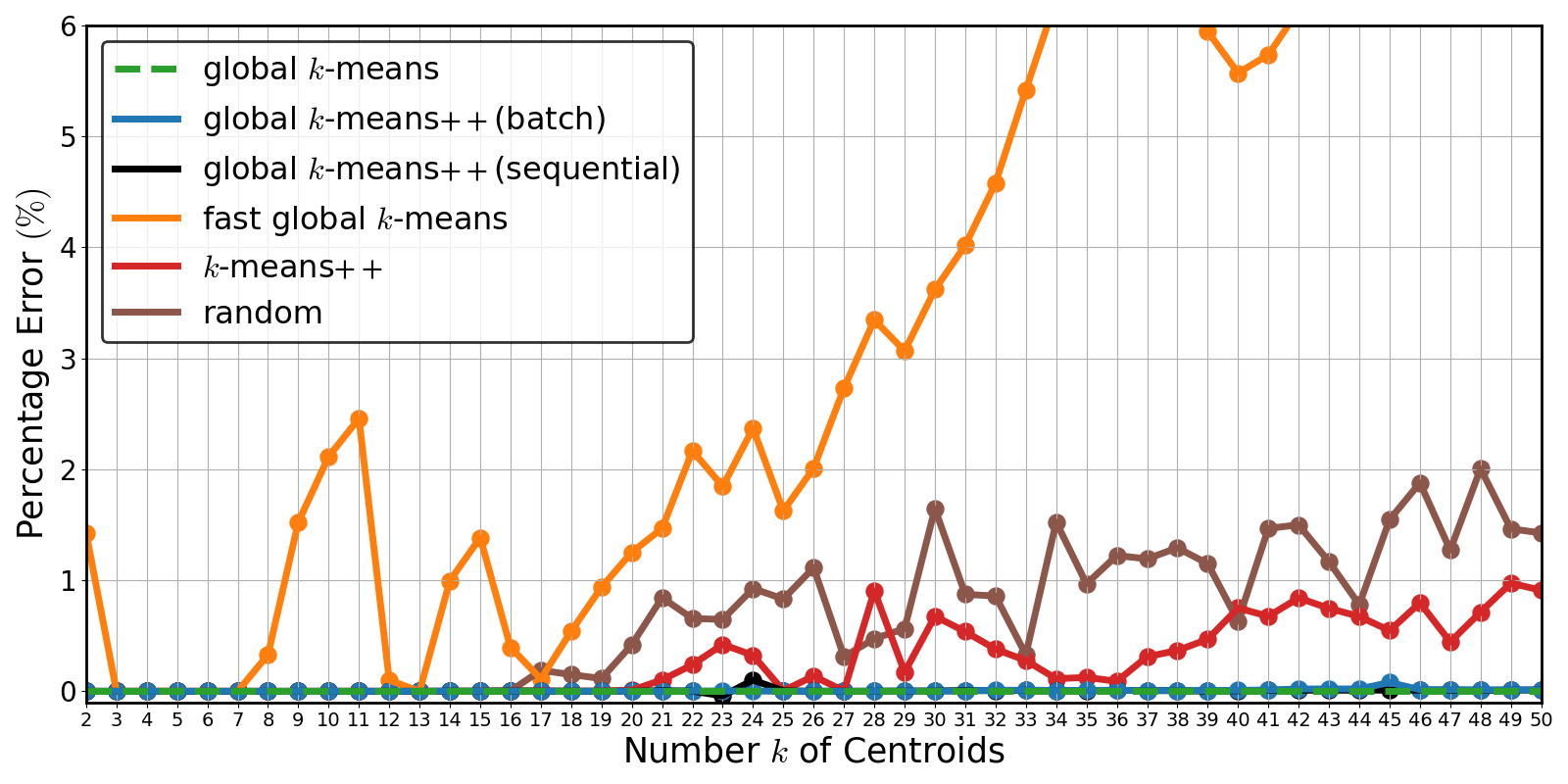"}
        \caption{}
        \label{}
    \end{subfigure}
    \caption{Relative Percentage Error for the Pendigits, for different $L$ values. (a) $L=10$ (b) $L=25$ (c) $L=50$ (d) $L=100$.}    
    \label{fig:Pendigits}
\end{figure}
	
\begin{figure}[t]
    \centering 
    \begin{subfigure}[b]{0.49\textwidth}
        \centering
        \includegraphics[width=\textwidth]{"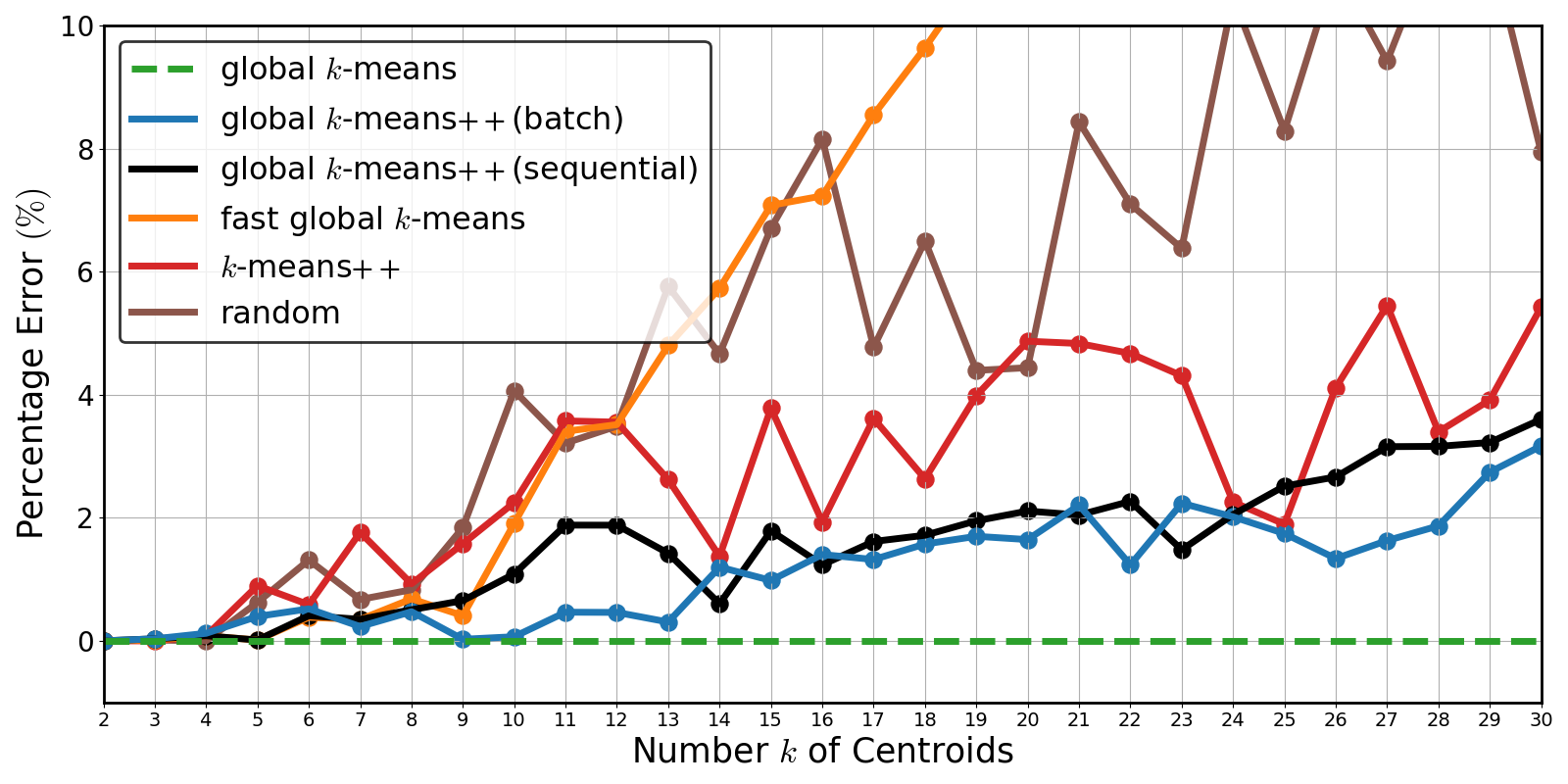"}
        \caption{}
        \label{}
    \end{subfigure}
    \hspace{0em}
    \begin{subfigure}[b]{0.49\textwidth}
        \centering
        \includegraphics[width=\textwidth]{"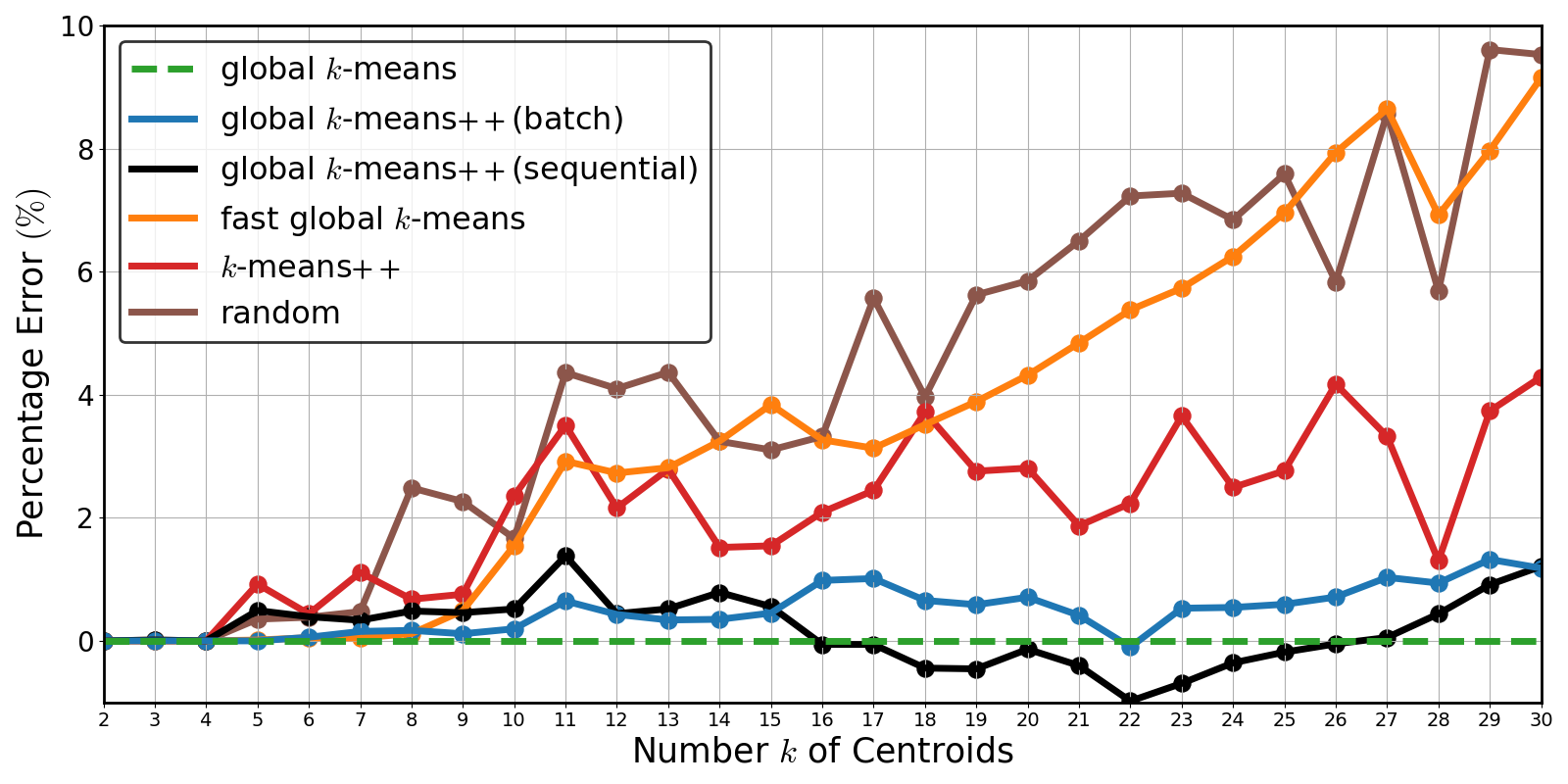"}
        \caption{}
        \label{}
    \end{subfigure}
    \begin{subfigure}[b]{0.49\textwidth}
        \centering
        \includegraphics[width=\textwidth]{"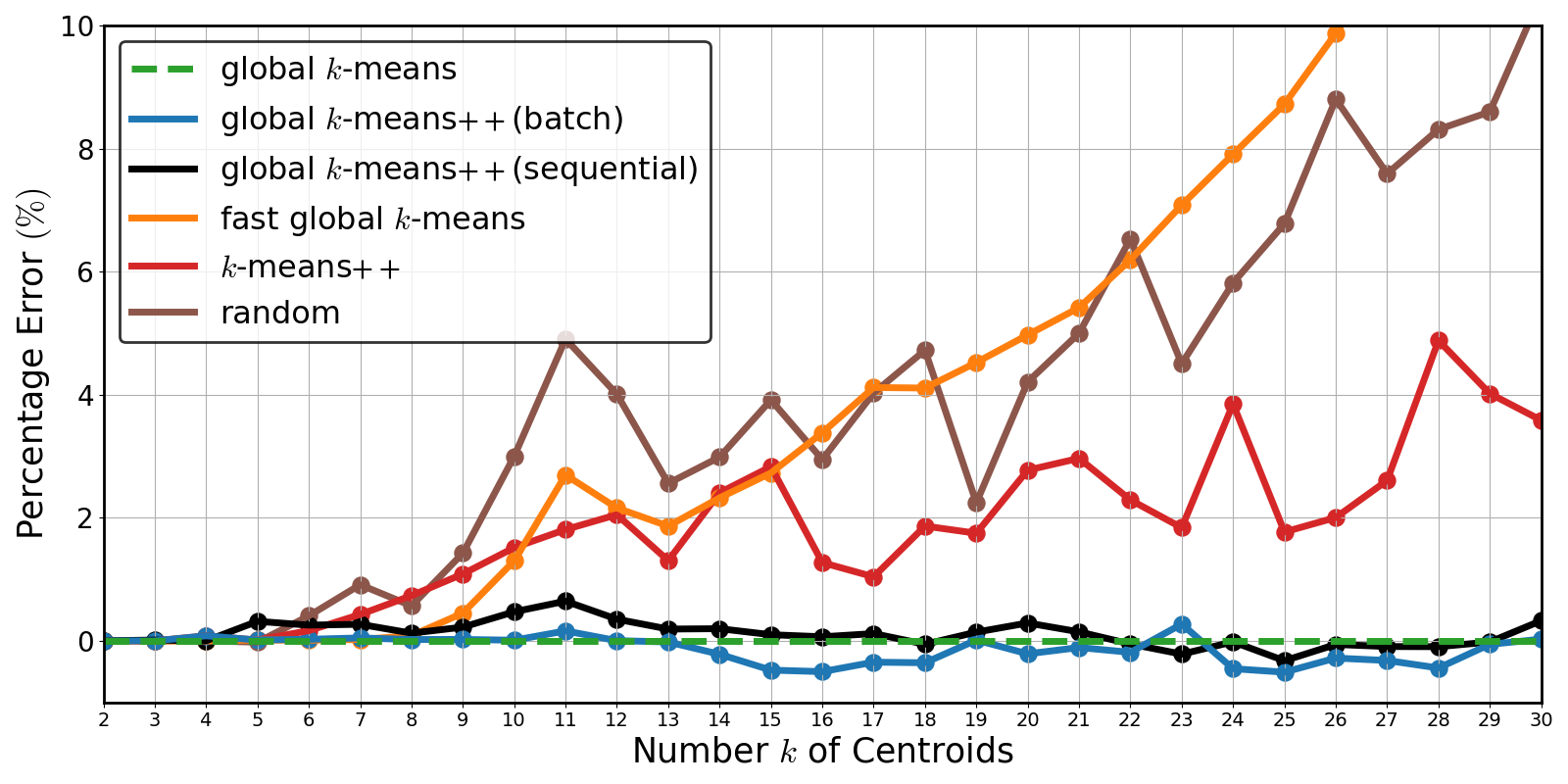"}
        \caption{}
        \label{}
    \end{subfigure}
    \hspace{0em}
    \begin{subfigure}[b]{0.49\textwidth}
        \centering
        \includegraphics[width=\textwidth]{"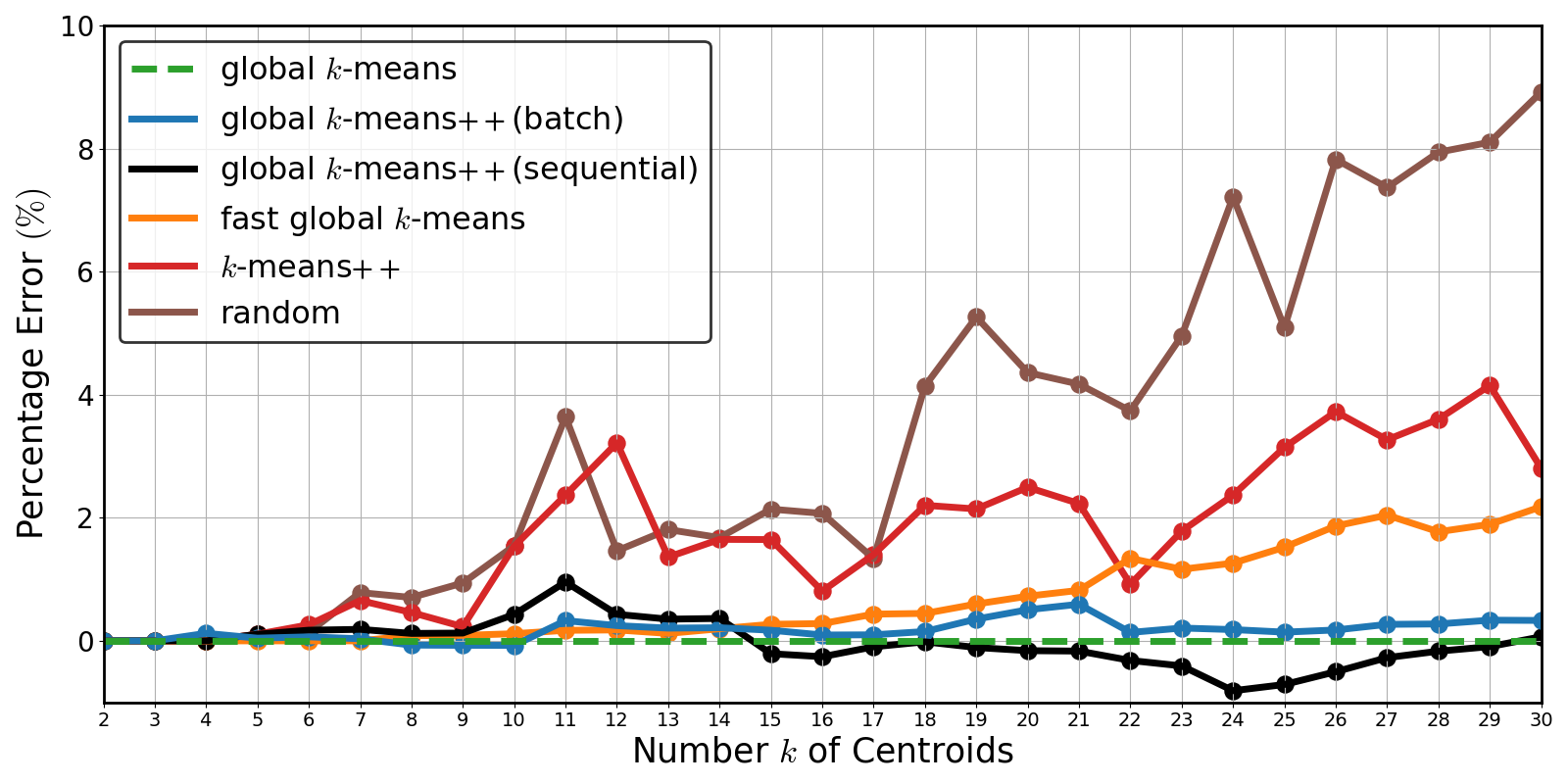"}
        \caption{}
        \label{}
    \end{subfigure}
    \caption{Relative Percentage Error for the Wine, for different $L$ values. (a) $L=10$ (b) $L=25$ (c) $L=50$ (d) $L=100$.}    
    \label{fig:Wine}
\end{figure}

\begin{figure}[t]
    \centering 
    \begin{subfigure}[b]{0.49\textwidth}
        \centering
        \includegraphics[width=\textwidth]{"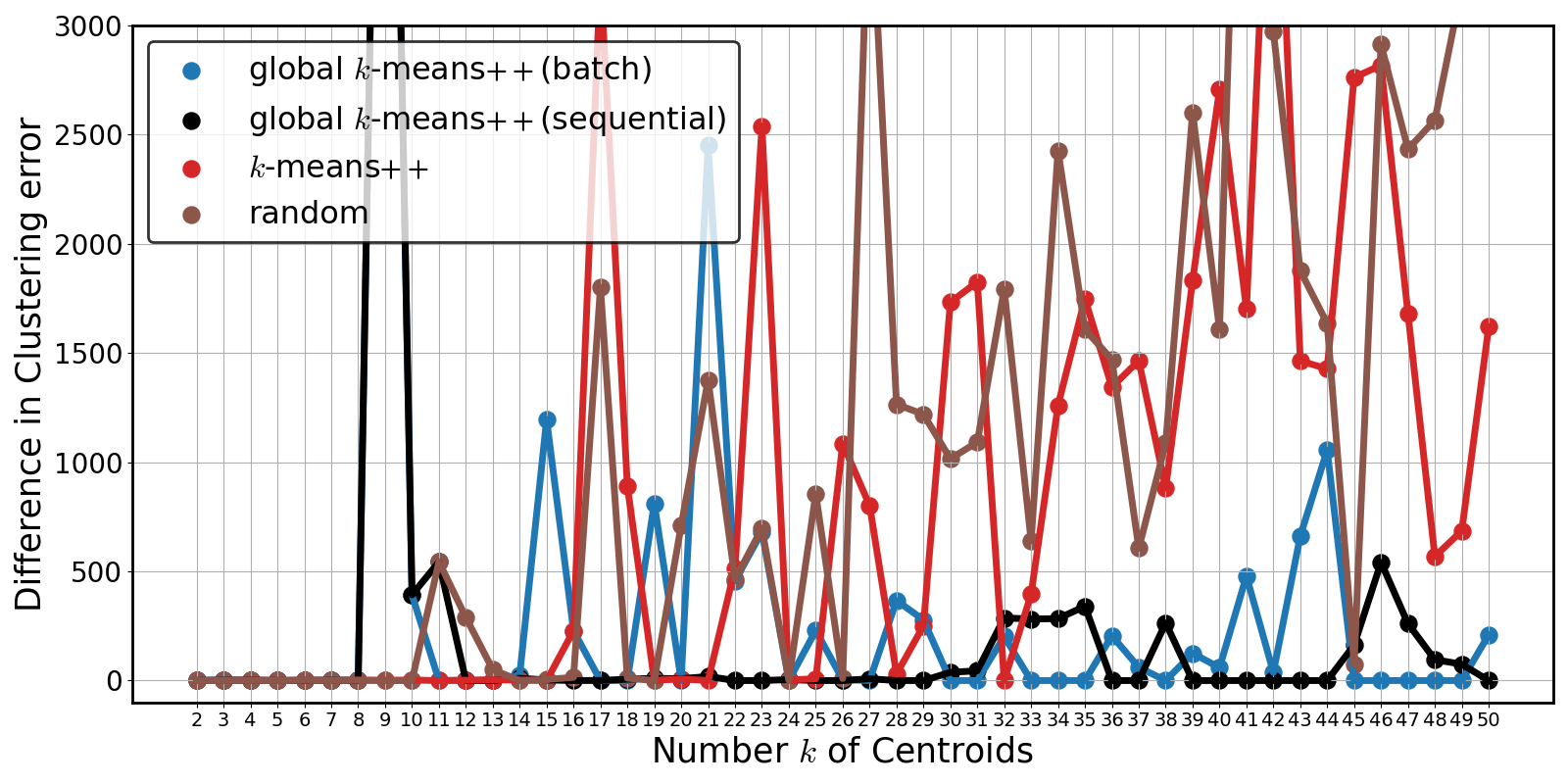"}
        \caption{}
        \label{}
    \end{subfigure}
    \hspace{0em}
    \begin{subfigure}[b]{0.49\textwidth}
        \centering
        \includegraphics[width=\textwidth]{"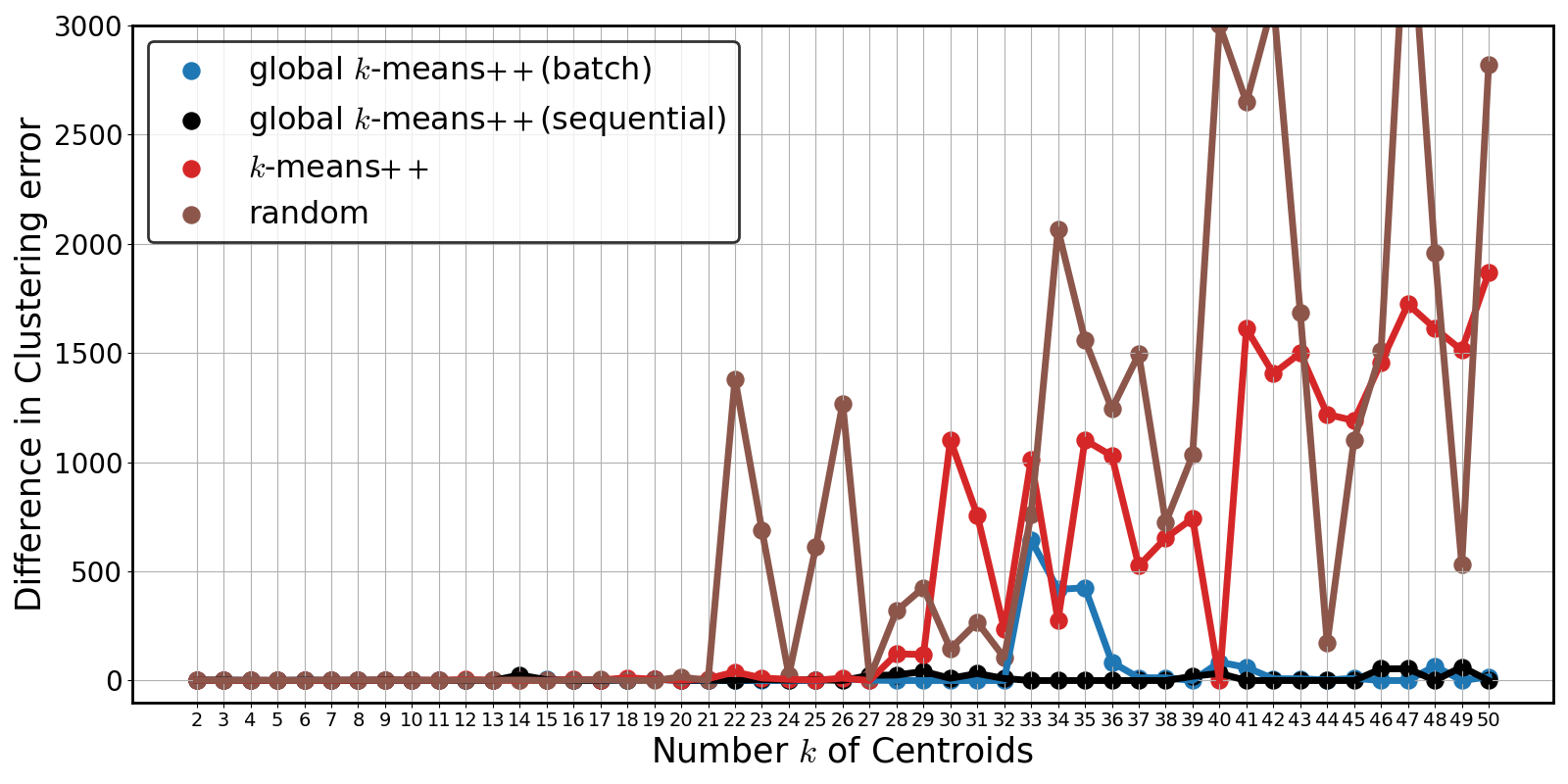"}
        \caption{}
        \label{}
    \end{subfigure}
    \begin{subfigure}[b]{0.49\textwidth}
        \centering
        \includegraphics[width=\textwidth]{"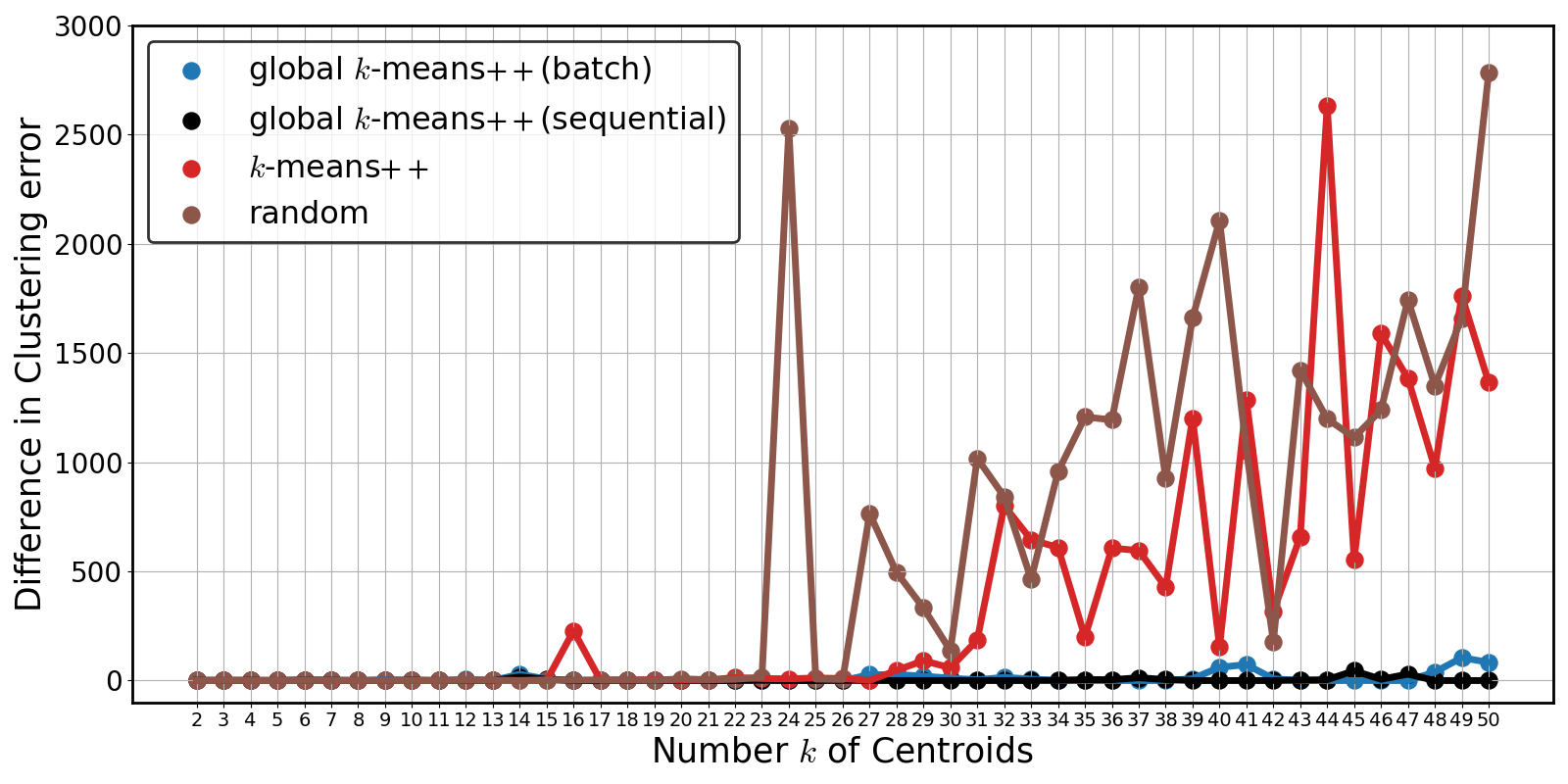"}
        \caption{}
        \label{}
    \end{subfigure}
    \hspace{0em}
    \begin{subfigure}[b]{0.49\textwidth}
        \centering
        \includegraphics[width=\textwidth]{"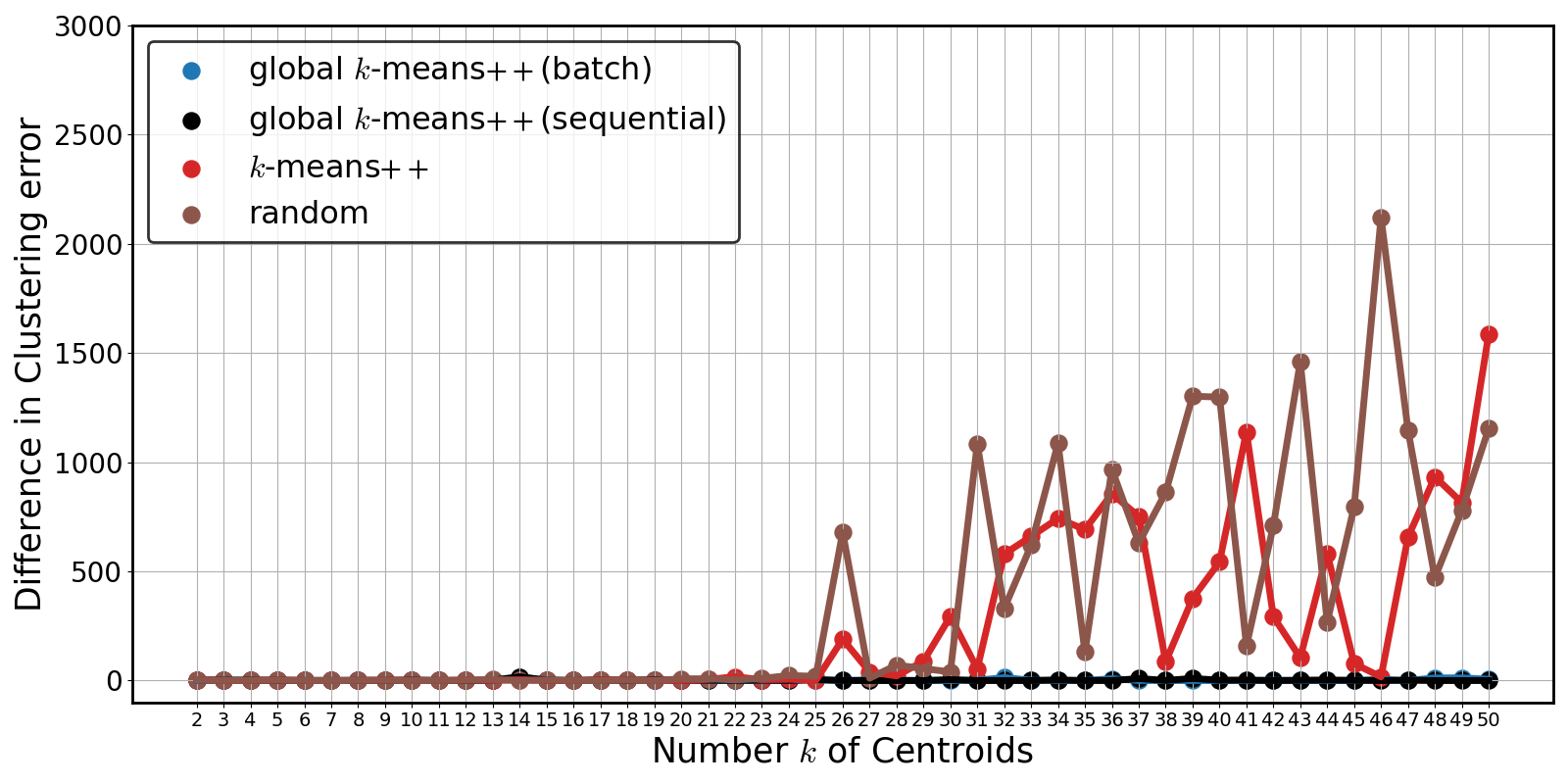"}
        \caption{}
        \label{}
    \end{subfigure}
    \caption{Clustering Error Differences for the Mnist, for different $L$ values. (a) $L=10$ (b) $L=25$ (c) $L=50$ (d) $L=100$.}    
    \label{fig:Mnist}
\end{figure}

\begin{figure}[t]
    \centering 
    \begin{subfigure}[b]{0.49\textwidth}
        \centering
        \includegraphics[width=\textwidth]{"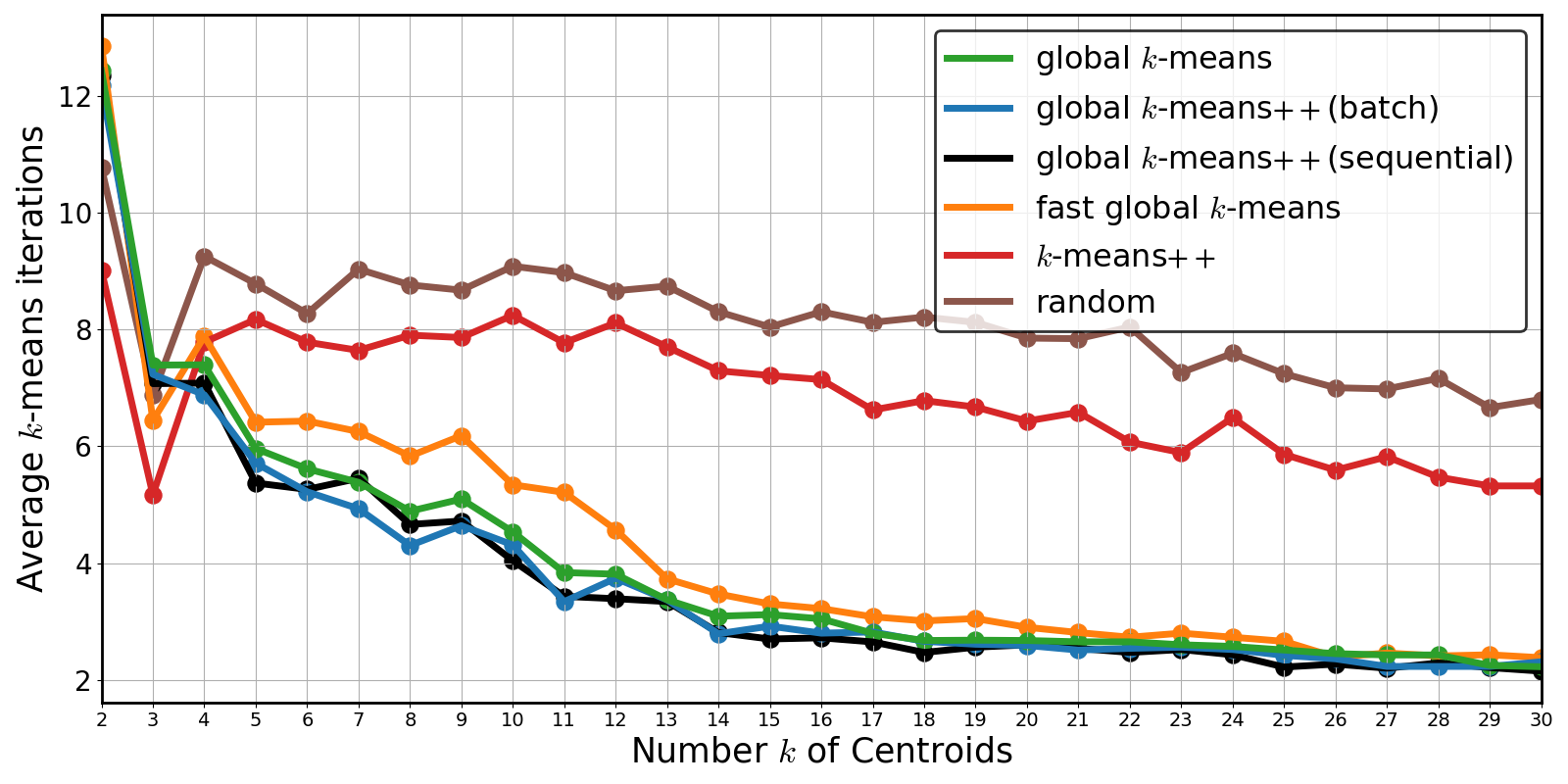"}
        \caption{}
        \label{}
    \end{subfigure}
    \hspace{0em}
    \begin{subfigure}[b]{0.49\textwidth}
        \centering
        \includegraphics[width=\textwidth]{"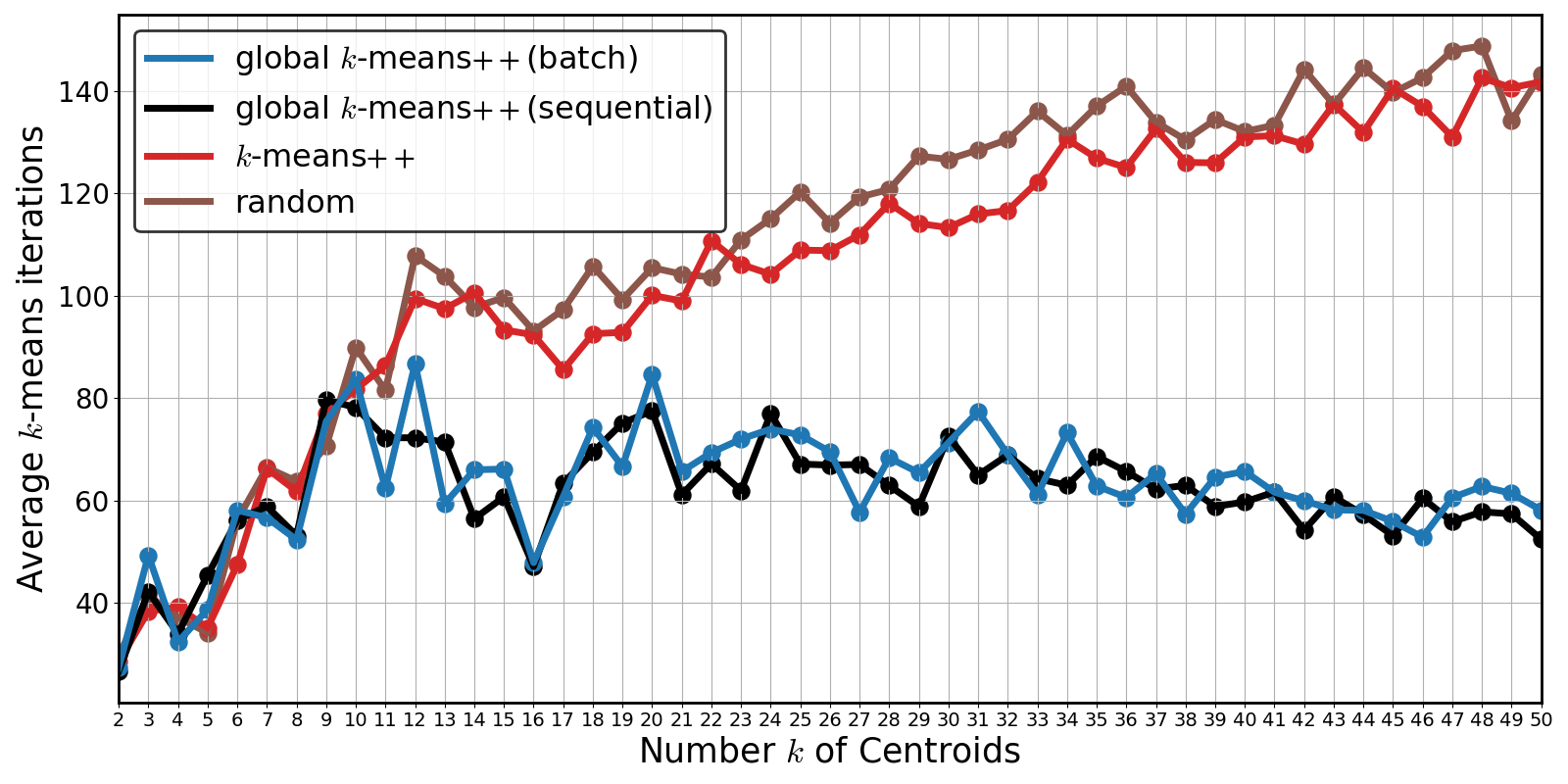"}
        \caption{}
        \label{}
    \end{subfigure}
    \begin{subfigure}[b]{0.49\textwidth}
        \centering
        \includegraphics[width=\textwidth]{"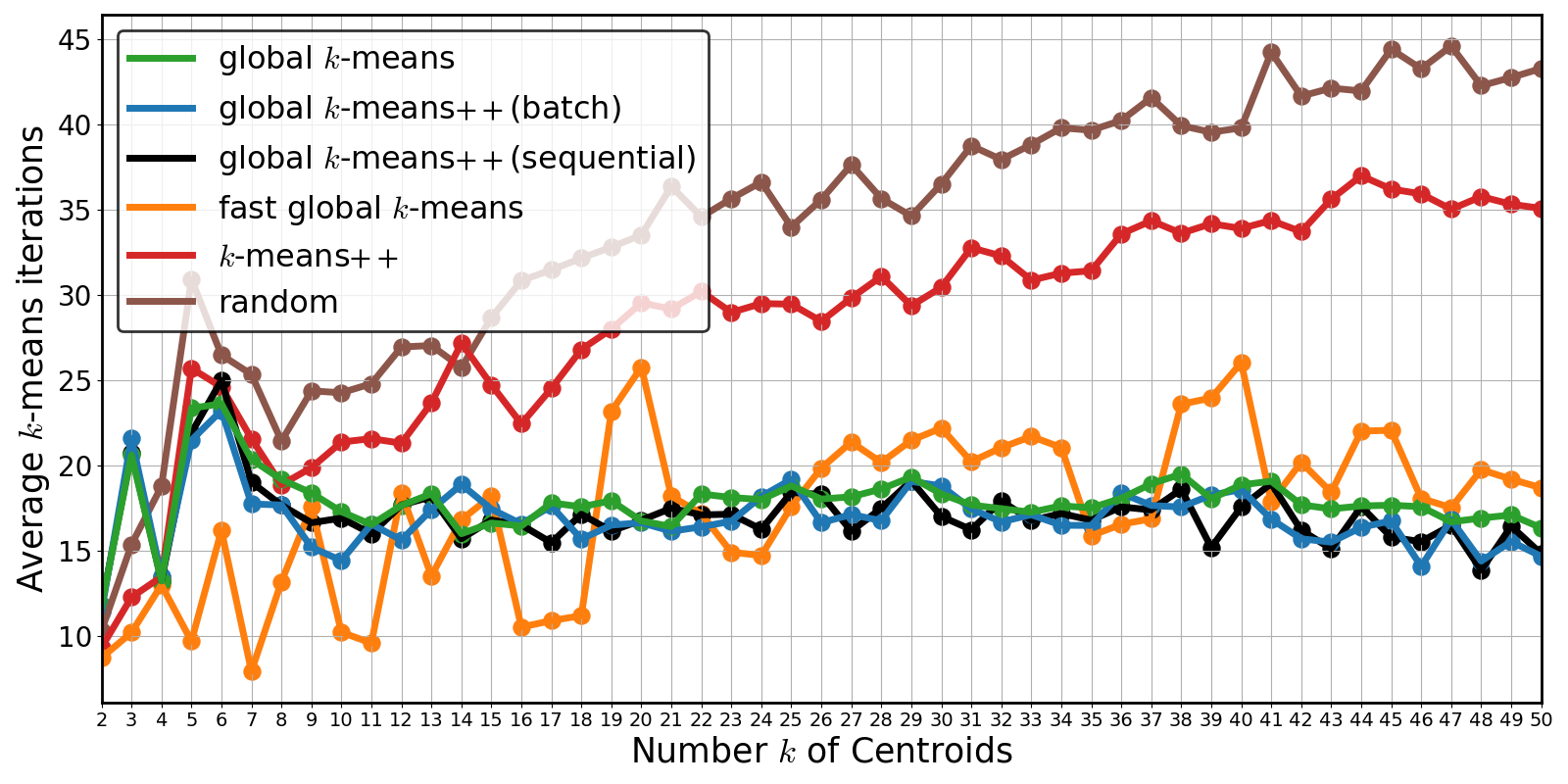"}
        \caption{}
        \label{}
    \end{subfigure}
    \hspace{0em}
    \begin{subfigure}[b]{0.49\textwidth}
        \centering
        \includegraphics[width=\textwidth]{"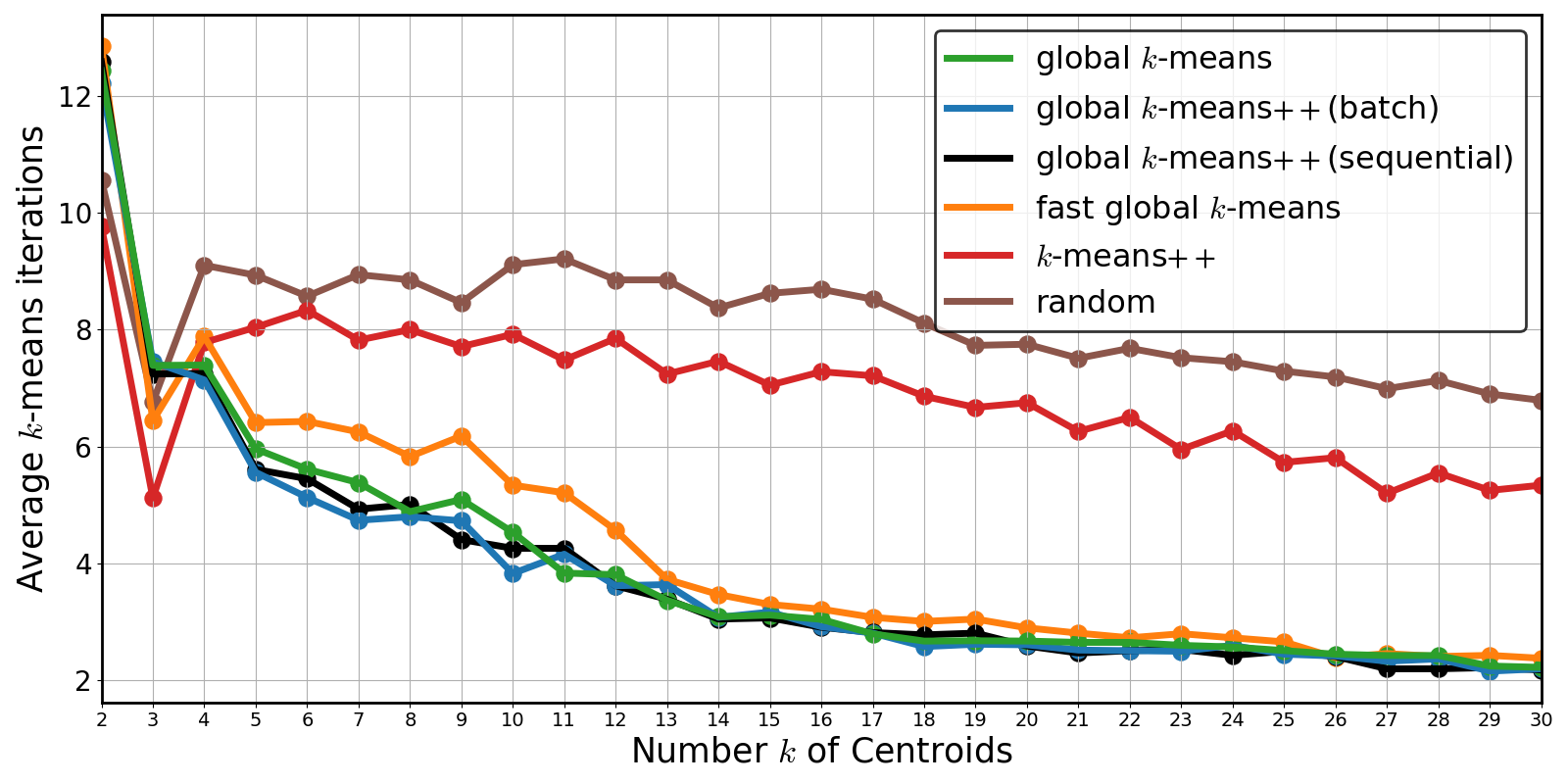"}
        \caption{}
        \label{}
    \end{subfigure}
    \caption{Average number of $k$-means iterations. (a) Breast Cancer (b) Mnist (c) Pendigits (d) Wine.}    
    \label{fig:avg_n_iter}
\end{figure}

\begin{table}[h]
	\centering
	\caption{CPU Time. Values marked by $\dagger$ and $\ddagger$ denote that the method could not be executed due to memory constraints or aborted after 7 days of execution respectively.}
	\label{tab:CPU_Time}
	\begin{tabular}{l@{\quad}r@{\quad}r@{\quad}r@{\quad}r@{\quad}r@{\quad}r@{\quad}r@{\quad}}
		\hlineb{1.5pt}
		Dataset & $L$ & gl & gl\texttt{++} (b) & gl\texttt{++} (s) & fgl & $k$-ms\texttt{++} & rnd \\
		\hlineb{1.5pt}
        
        \multirow{4}{*}{Breast Cancer} & 10 & \multirow{4}{*}{32.67s} & \textbf{0.48s} & \underline{0.64s} & 0.88s & 1.57s & 1.18s \\
        & 25 & & \textbf{1.30s} & \underline{1.52s} & 1.80s & 4.10s & 3.04s \\
        & 50 & & \textbf{2.55s} & \underline{3.12s} & 3.40s & 8.06s & 6.40s \\
        & 100 & & \textbf{5.46s} & \underline{5.86s} & 6.21s & 15.03s & 12.31s \\
        
        \hline
        \multirow{4}{*}{Mnist} & 10 & \multirow{4}{*}{$\dagger$} & \textbf{8.35m} & \underline{9.4m} & \multirow{4}{*}{$\ddagger$} & 29.81m & 15.5m \\
         & 25 & & \textbf{25.58m} & \underline{30.23m} &  & 1.77h & 39.73m \\
         & 50 & & \textbf{49.5m} & \underline{57.25m} & & 3.68h & 1.39h \\
         & 100 & & \textbf{1.58h} & \underline{1.81h} & & 6.7h & 2.70h \\

        \hline
        \multirow{4}{*}{Pendigits} & 10 & \multirow{4}{*}{4d} & \textbf{5.58m} & \underline{5.77m} & 7.25m & 12.92m & 13.27m \\
        & 25 & & \textbf{14m} & \underline{14.97m} & 16.9m & 31.87m & 32.97m \\
        & 50 & & \underline{28.42m} & \textbf{28.17m} & 35.48m & 1.09h & 1.12h \\
        & 100 & & \textbf{55.07m} & \underline{1.09h} & 1.16h & 2.2h & 2.2h \\

        \hline
    	\multirow{4}{*}{Wine} & 10 & \multirow{4}{*}{5.40s} & \textbf{0.32s} & 0.40s & \underline{0.36s} & 1.10s & 0.49s \\
        & 25 & & \textbf{0.76s} & 0.95s & \underline{0.84s} & 2.77s & 1.21s \\
        & 50 & & \textbf{1.51s} & 1.90s & \underline{1.66s} & 5.47s & 2.46s \\
        & 100 & & \textbf{3.06s} & 3.81s & \underline{3.24s} & 10.77s & 5.10s \\
		\hlineb{1.5pt}
	\end{tabular}
\end{table}

Figures \ref{fig:Breast-Cancer}, \ref{fig:Pendigits}, and \ref{fig:Wine} depict the relative percentage error between each method and the baseline algorithm, for the respective datasets: Breast Cancer, Pendigits, and Wine. However, in the case of the Mnist dataset, Figure~\ref{fig:Mnist} displays the difference in clustering error between each method and the best-performing algorithm. In the subfigures (a), (b), (c), and (d), we present the results of experiments considering different indicative values of $L$ (i.e., 10, 25, 50, and 100) for the global $k$-means variants, $k$-means\texttt{++}, and the standard $k$-means algorithm. Moreover, Figure~\ref{fig:avg_n_iter} provides the average number of iterations required by each algorithm to converge, while Table~\ref{tab:CPU_Time} presents the CPU time necessary to compute all $K$ clustering solutions for each algorithm.

The results across all datasets demonstrate that both the batch and sequential versions of global $k$-means\texttt{++} exhibit similar performance to global $k$-means, while clearly outperforming the FGKM, $k$-means\texttt{++}, and the standard $k$-means algorithm. The optimization capabilities of global $k$-means\texttt{++} are particularly noteworthy, especially as $k$ increases, where minimizing the clustering error becomes more challenging. As expected, the performance of global $k$-means\texttt{++} improves with an increasing number of candidates, better approximating the performance of global $k$-means. Surprisingly, in the Wine dataset, global $k$-means\texttt{++} outperformed global $k$-means in several $k$ sub-problems (Fig. \ref{fig:Wine}b-\ref{fig:Wine}d). The relative percentage error of global $k$-means\texttt{++} was less than $1\%$ when considering more than $L=25$ candidates. In the Pendigits dataset~\ref{fig:Pendigits}, the baseline model and our algirithm coincide with high accuracy for $L=25, 50, 100$. Notably, global $k$-means required four days of computation, while global $k$-means\texttt{++} completed the task in significantly less time, ranging from 14 minutes to 1 hour, depending on the value of $L$. This indicates that we achieved comparable results in a fraction of the time. Last but not least, in the case of the Mnist dataset~\ref{fig:Mnist}, when we consider $L=25$ candidates or more, the global $k$-means\texttt{++} consistently outperforms the other methods.

It should be noted that in the Wine dataset (\ref{fig:Wine}d), the FGKM algorithm with $L=100$ candidates demonstrated promising results. However, its performance across all datasets was inconsistent. It is evident that the FGKM algorithm cannot reliably approximate the clustering solution achieved by the global $k$-means algorithm. Even with an increasing number of candidates, the clustering solution provided by the FGKM significantly deviates from the quality attained by the global $k$-means\texttt{++}. As expected, the $k$-means\texttt{++} algorithm was successful for small values of $k$ across all datasets. However, it becomes apparent that as $k$ increases, the performance of the $k$-means\texttt{++} algorithm deteriorates compared to global $k$-means and global $k$-means\texttt{++}. This is because the $k$-means problem becomes more challenging when it comes to selecting the initial center positions as $k$ raises. As expected, the standard $k$-means with the random uniform selection yielded the worst results. 

Figure~\ref{fig:avg_n_iter} presents the average number of $k$-means iterations performed by each method in all datasets and $k$ sub-problems. It becomes evident that as $k$ increases, the incremental methods require fewer iterations in each run of $k$-means due to improved initialization of the cluster centers. This advantage arises from the fact that the previous $k-1$ centers have already been positioned in near-optimal solutions. It should be reminded that for each cluster number $k$, the number of executed $k$-means runs is $\mathcal{O}(Lk)$ for all compared methods, except for global $k$-means which executes $k$-means $\mathcal{O}(Nk)$ times. This means that even if global $k$-means tends to execute a smaller average number of $k$-means iterations compared to $k$-means\texttt{++}, this does not translate to faster execution times. Table~\ref{tab:CPU_Time} presents the CPU time required by each algorithm to compute all $K$ clustering solutions. It turns out that if we want to find all $k$ cluster solutions from $k=2$ until a cluster number $K$, global $k$-means\texttt{++} is the faster approach. 
  
\section{Conclusions}
\label{sec:Conclusion}
We have proposed the global $k$-means\texttt{++} clustering algorithm, which is an effective relaxation of the global $k$-means algorithm that provides an ideal compromise between clustering error and execution speed. The basic idea of the proposed method is to take advantage of the superior clustering solutions that the global $k$-means algorithm can provide while avoiding its substantial computational requirements. The global $k$-means\texttt{++} is an incremental clustering approach that dynamically adds one cluster center at each $k$ cluster sub-problem. For each $k$ cluster sub-problem, the method selects $L$ datapoints as candidates for the initial position of the new center using the effective $k$-means\texttt{++} selection probability distribution. The selection method is fast and requires no extra computational effort for distance computations. 
	
Global $k$-means\texttt{++} has been tested on various benchmark publicly available datasets and has been compared to the global $k$-means, the FGKM with multiple candidates, the $k$-means\texttt{++} and the standard $k$-means with random uniform initialization. The experimental results reveal its superiority against the FGKM, the $k$-means\texttt{++}, and the standard $k$-means algorithms. Furthermore, its performance is comparable to the global $k$-means with a significantly reduced computational cost, thus establishing it as an effective relaxation.
 
We conclude that the proposed algorithm balances out the best characteristics of global $k$-means and $k$-means\texttt{++} methods, resulting in high quality clustering at reduced computational cost. Despite previous efforts to enhance the FGKM algorithm, our method takes a different approach by directly addressing the global $k$-means method. We believe this method is an effective alternative to both global k-means and k-means++. Furthermore, the method provides all clustering solutions for $k\in \{1,\ldots, K\}$. In this way, it is suitable for comparing clustering solutions for different $k$ to determine the appropriate number of clusters, which is unknown in many applications. In this case, the method is significantly faster, even when compared to non-incremental methods such as standard $k$-means and $k$-means\texttt{++}.
	
In future work, it would be interesting to investigate the global $k$-means\texttt{++} algorithm into a semi-supervised setting by incorporating must-link or cannot-link constraints by utilizing the exact solver as referred to~\cite{piccialli2022exact}. In addition, we also aim to test the method in real applications (such as in biology, speech recognition, face clustering, etc.) initially requiring a large number of clusters (overclustering solutions). Finally, it would be interesting to integrate the method into a deep framework for clustering of composite objects (graphs, images, text, etc.).
    
\section*{Acknowledgements}
This research was supported by project ``Dioni: Computing Infrastructure for Big-Data Processing and Analysis'' (MIS No. 5047222) co-funded by European Union (ERDF) and Greece through Operational Program “Competitiveness, Entrepreneurship and Innovation”, NSRF 2014-2020.

\subsection*{Data Availability Statement}
\noindent$\bullet$ The \textbf{UCI} datasets are available at the official UCI page: \href{https://archive.ics.uci.edu}{https://archive.ics.uci.edu/ml/index.php}. 

\noindent$\bullet$ The \textbf{MNIST} dataset is available at the following official website: \href{http://yann.lecun.com/exdb/mnist/}{http://yann.lecun.com/exdb/mnist/}.
	
\bibliographystyle{unsrt}
\bibliography{Bibliography.bib}

\end{document}